\documentclass[10pt,twocolumn,letterpaper]{article}

\usepackage{wacv}
\usepackage{times}
\usepackage{epsfig}
\usepackage{graphicx}
\usepackage{amsmath}
\usepackage{amssymb}

\usepackage[flushleft]{threeparttable}
\usepackage{balance}

\usepackage{booktabs}
\usepackage{footmisc}
\usepackage{caption}
\usepackage{subcaption}

\wacvfinalcopy 

\ifwacvfinal
\def\assignedStartPage{1} 
\fi

\ifwacvfinal
\usepackage[breaklinks=true,bookmarks=false]{hyperref}
\else
\usepackage[pagebackref=true,breaklinks=true,colorlinks,bookmarks=false]{hyperref}
\fi

\ifwacvfinal
\else
\fi

\begin{document}

\title{ENIGMA-51: Towards a Fine-Grained Understanding of Human-Object Interactions in Industrial Scenarios}

\author{Francesco Ragusa $^{1,2}$, Rosario Leonardi$^{1}$, Michele Mazzamuto$^{1,2}$,\\ Claudia Bonanno$^{1,2}$, Rosario Scavo$^{1}$, Antonino Furnari$^{1,2}$, Giovanni Maria Farinella$^{1,2}$\\
\\
$^{1}$FPV@IPLab - University of Catania, Italy\\
$^{2}$Next Vision s.r.l. - Spinoff of the University of Catania, Italy
\\
{\tt\small }
}
\maketitle

\begin{abstract}
ENIGMA-51 is a new egocentric dataset acquired in an industrial scenario by 19 subjects who followed instructions to complete the repair of electrical boards using industrial tools (e.g., electric screwdriver) and equipments (e.g., oscilloscope). The 51 egocentric video sequences are densely annotated with a rich set of labels that enable the systematic study of human behavior in the industrial domain. We provide benchmarks on four tasks related to human behavior: 1) untrimmed temporal detection of human-object interactions, 2) egocentric human-object interaction detection, 3) short-term object interaction anticipation and 4) natural language understanding of intents and entities. Baseline results show that the ENIGMA-51 dataset poses a challenging benchmark to study human behavior in industrial scenarios. We publicly release the dataset at  \url{https://iplab.dmi.unict.it/ENIGMA-51}.
\end{abstract}

\section{Introduction}
Every day, humans interact with the surrounding world to achieve their goals. 
These interactions are often complex and require multiple steps, skills, and involve different objects. 
For example, in an industrial workplace, when performing maintenance of industrial machinery, a worker interacts with several objects and tools while repairing the machine (e.g., \textit{wear PPEs, take the screwdriver}), testing it (e.g., \textit{press the button on the electric panel}), and writing a report (e.g., \textit{take the pen, write the report}).
To properly assist humans, an intelligent system should be able to model human-object interactions (HOIs) from real-world observations captured by users wearing smart cameras (e.g., smart glasses)~\cite{Colombo3316782.3322754,DANIELSSON20201298,mazzamuto2023wearable}. It is also plausible that predicting human-object interactions in advance can benefit an intelligent system help workers to avoid mistakes, or to improve their safety. For example, during the execution of a maintenance procedure, an AI assistant should be able to understand when the worker is interacting with the objects, show technical information, provide instructions on how to interact with each object, alert the worker of potential safety risks (e.g., \textit{Before touching the electrical board, turn off the power supply!}), and suggest what the next interaction is. Furthermore, an intelligent system should be able to have a natural language conversation with workers. It should also be able to extract useful information from their speech, and figure out what they are trying to achieve. This way, it can provide assistance for supporting their needs, preferences, and goals.

In general, tasks focused on understanding human behaviour have been extensively studied thanks to the availability of public datasets that consider multiple domains~\cite{Grauman2022Ego4DAT, Hands_in_contact_Shan20, EgoProceLECCV2022, Liu_2022_CVPR} or specific ones, such as kitchens~\cite{damen2018epickitchen, lu2021egocentric, zhukov2019crosstask}, daily life~\cite{Pirsiavash2012DetectingAO, Lee_UTE_2012}, and industrial-like scenarios~\cite{sener2022assembly101,ragusa2023meccano}.
However, since data acquisition in a real industrial scenario is challenging due to privacy issues, safety and industrial secret protection, the datasets available to date do not reflect real industrial environments, considering proxy activities such as employing toy models made of textureless parts~\cite{ragusa2023meccano, sener2022assembly101}.

Considering what stated above, to enable research in this field, we present ENIGMA-51, a new dataset composed of 51 egocentric videos acquired in an industrial environment which simulates a real industrial laboratory. The dataset was acquired by 19 subjects who wore a Microsoft HoloLens 2~\cite{Holo2} headset and followed audio and AR instructions provided by the device to complete repairing procedures on electrical boards. The subjects interact with industrial tools such as an electric screwdriver and pliers, as well as with electronic instruments such as a power supply and an oscilloscope while executing the steps to complete a specific procedure. \begin{figure*}[ht]
    \centering
    \includegraphics[width=0.9\linewidth]{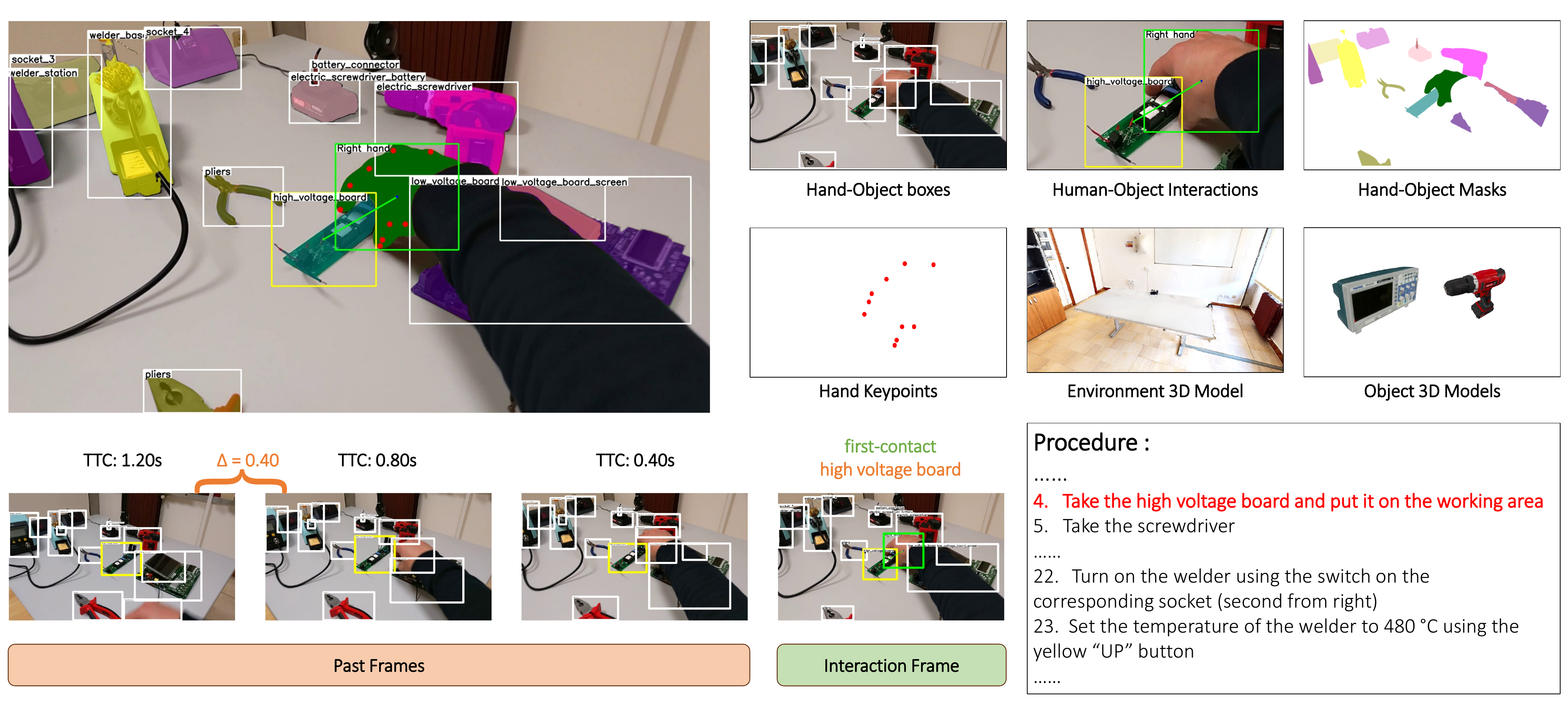}
    \caption{Frames have been annotated with a rich set of labels (top-left). Sequences have been annotated by determining the interaction key frame (bottom-center), assigning the verb (green) and the active object (orange). For each interaction key frame, we provide objects and hand bounding boxes and the relation between them. In the past frames, we annotated also the next active objects and we derived the time to contact (TTC) (bottom-left). We also generated pseudo-labels for semantic masks and hand keypoints, and we released 3D models for the objects and for the laboratory (top-right). Moreover, a specific instruction belonging to the procedure is associated with each interaction key frame (bottom-right).}
    \label{fig:labels}
\end{figure*}

Apart the current interactions, we annotated which objects and hands will be involved in future interactions, as well as the time to contact (TTC) to indicate when the future interaction will start. This allows us to explore the task of predicting the future human-object interactions considering the industrial domain. Textual instructions used for the data acquisition, also allow to study tasks which focus on the knowledge extraction of intents and entities from the text while users are interacting with the objects. In the industrial domain these tasks have not been explored due to the lack of public egocentric datasets explicitly annotated with intents and entities.

Together with the manually annotations, we release the pseudo-labels and the pre-extracted features to enable further investigations beyond the current study. In particular, we generated hands and objects segmentation masks~\cite{sam_hq}, and hands keypoints~\cite{mmpose2020}. The provided visual features are extracted with DINOv2~\cite{oquab2023dinov2} and CLIP~\cite{Radford2021LearningTV}. To allow further research in the context of scalable models trained using synthetic data, we share the 3D models of the laboratory and all considered industrial objects. Figure~\ref{fig:labels} shows examples of images acquired in the industrial environment where the dataset was acquired together with the annotations. To highlight the usefulness of the proposed dataset, we performed baseline experiments related to 4 fundamental tasks focused on understanding human behavior from first person vision in the considered industrial context: 1) Untrimmed Temporal Detection of Human-Object Interactions, 2) Egocentric Human-Object Interaction (EHOI) Detection, 3) Short-Term Object Interaction Anticipation and 4) Natural Language Understanding of Intents and Entities. 

In sum the contributions of this work are as follows: 1) we introduce ENIGMA-51, a new dataset composed of 51 egocentric videos acquired in an industrial domain; 2) we manually annotated the dataset with a rich set of annotations aimed at studying human behavior; 3) we propose a benchmark to study human behavior in an industrial environment exploring 4 different tasks, showing that the current state-of-the-art approaches are not sufficient to solve the considered problems in the industrial setting; 4) we provide additional labels and features exploiting foundational models, with the aim to push research on additional tasks on the proposed industrial dataset.
The ENIGMA-51 dataset and its annotations are available at the following link: \url{https://iplab.dmi.unict.it/ENIGMA-51}.

\section{Related Work}
Our work is related to previous research lines which are revised in the following sections.

\subsection{Ego Datasets for Human Behavior Studies}

\begin{table*}[t]
\centering
\resizebox{0.8\textwidth}{!}{%
\begin{tabular}{llcccccc}
\textbf{Dataset} & \multicolumn{1}{c}{\textbf{Year}} & \multicolumn{1}{l}{\textbf{Video?}} & \multicolumn{1}{l}{\textbf{EHOI Annotations?}} & \textbf{Settings} & \textbf{Hours} & \textbf{Sequences} & \textbf{Subjects} \\ \hline
ENIGMA-51 (ours) & 2024 & \checkmark & \checkmark & Industrial & 22 & 51 & 19 \\
MECCANO\cite{ragusa2023meccano} & 2023 & \checkmark & \checkmark & Industrial-like & 7 & 20 & 20 \\ \hline
Ego4D\cite{Grauman2022Ego4DAT} & 2022 & \checkmark & \checkmark & Multi-domain & 3670 & 9650 & 923 \\
THU-READ\cite{thu-read_Tang_2017} & 2019 & \checkmark & \checkmark & Daily activities & 224 & 1920 & 8 \\
EPIC-KITCHENS-VISOR\cite{VISOR2022} & 2022 & \checkmark & \checkmark & Kitchen activities & 100 & 700 & 45 \\
HOI4D\cite{Liu_2022_CVPR} & 2022 & \checkmark & \checkmark & Objects manipulation & 22 & 4000 & N/A \\
VOST\cite{tokmakov2023breaking} & 2023 & \checkmark & \checkmark & Daily + Industrial-like & 4 & 713 & N/A \\
ARCTIC\cite{fan2023arctic} & 2023 & \checkmark & \checkmark & Object manipulation & 2 & 339 & 10 \\
\hline
100 Days of Hands\cite{Hands_in_contact_Shan20} & 2020 & X & \checkmark & Daily activities & 3144 & 27000 & 1350+ \\
GUN-71\cite{gun-71_dataset_Rogez_2015} & 2015 & X & \checkmark & Daily activities & N/A & N/A & 8 \\ \hline
Assembly101\cite{sener2022assembly101} & \multicolumn{1}{c}{2022} & \checkmark & X & Industrial-like & 513 & 362 & 53 \\
EGTEA Gaze+\cite{li2020eye} & 2017 &\checkmark & X & Cooking activities & 28 & 86 & 32 \\
ADL\cite{Pirsiavash2012DetectingAO} & 2012 & \checkmark & X & Daily activities & 10 & 20 & 20 \\ \hline
\end{tabular}
}
\caption{Overview of egocentric datasets with a particular focus on those that allow the study of human-object interactions sorted by the number of hours.}
\label{tab:ego-dataset}
\end{table*}

Previous works have proposed egocentric datasets focusing on human behavior understanding.
The Activity of Daily Living (ADL)~\cite{Pirsiavash2012DetectingAO} dataset is one of the first datasets acquired from the egocentric perspective. It includes 20 egocentric videos where participants were involved in daily activities. It comprises temporal action annotations aimed to study egocentric activities.
EGTEA Gaze+~\cite{li2020eye} focuses on cooking activities involving 32 subjects who recorded 28 hours of videos. It has been annotated with pixel-level hand masks and 10325 action annotations including 19 action verbs and 51 object nouns. The THU-READ~\cite{thu-read_Tang_2017} dataset is composed of 1920 RGB-D sequences captured by 8 participants who performed 40 different daily-life actions. The EPIC-Kitchens datasets~\cite{damen2018epickitchen, Damen2022RESCALING} are collections of egocentric videos that capture natural actions in kitchen settings. EPIC-Kitchens-55~\cite{damen2018epickitchen} consists of 432 videos with annotations for 352 objects and 125 verbs. EPIC-Kitchens-100~\cite{Damen2022RESCALING} is a larger version of EPIC-Kitchens-55 with more videos (700), scenes (45) and hours (100). Assembly101~\cite{sener2022assembly101} simulates an industrial scenario and it is composed of 4321 assembly and disassembly videos of toy vehicles made of textureless parts. It offers a multi-view perspective, comprising static and egocentric recordings annotated with 100K coarse and 1M fine-grained action segments and with 18M 3D hand poses.

While these datasets explore actions and activities, other datasets have been proposed to study human-object interactions from the egocentric perspective in a more fine-grained fashion.
The Grasp Understanding (GUN-71~\cite{gun-71_dataset_Rogez_2015}) dataset, contains 12,000 images of hands manipulating 28 objects labelled with 71 grasping categories. The 100 Days Of Hands (100DOH)~\cite{Shan20} dataset captures hands and objects involved in generic interactions. It consists of 100K frames collected over 131 days with 11 types of interactions. It comprises bounding boxes around the hands and the active objects, the side of the hands and the contact state (which indicates if the hand is touching an object or not). Other works focused on human-object interactions providing egocentric video datasets. 
EPIC-KITCHENS VISOR~\cite{VISOR2022} contains videos from EPIC-KITCHENS-100~\cite{Damen2022RESCALING} annotated with 272K semantic masks for 257 classes of objects, 9.9M interpolated dense masks, and 67K human-object interactions. The authors of~\cite{Liu_2022_CVPR} proposed the HOI4D dataset which is composed of 2.4 million RGB-D egocentric frames across 4000 sequences acquired in 610 indoor rooms. 
The authors of~\cite{fan2023arctic} studied hands interacting with articulated objects (e.g., scissors, laptops) releasing the ARCTIC dataset. It comprises 2.1M high-resolution images annotated with 3D hand and object meshes and with contact information. The VOST dataset~\cite{tokmakov2023breaking} focuses on objects that dramatically change their appearance. It includes 713 sequences where the objects have been annotated with segmentation masks. Ego4D~\cite{Grauman2022Ego4DAT} is a massive-scale dataset composed of 3670 hours of daily-life activity videos acquired in different domains by 923 unique participants. It comes with a rich set of annotations to address tasks concerning the understanding of the past, present, and future.

More related to our work are datasets acquired in the industrial-like domain~\cite{ragusa2023meccano, sener2022assembly101}. Unlike  Assembly101~\cite{sener2022assembly101} and MECCANO~\cite{ragusa2023meccano} we consider an industrial setting which simulates a real industrial laboratory. Unlike Assembly101, we provide fine-grained annotations to study different aspects of human behavior.

Table~\ref{tab:ego-dataset} shows the key attributes of the analyzed datasets. Previous datasets have focused on kitchens, daily activities, and industrial-like scenarios exploring different aspects of the human behavior. In order to perform a systematic study on human behaviour and human-object interactions in an industrial domain, we present the ENIGMA-51 dataset with a rich set of fine-grained egocentric videos together with annotations.

\subsection{Untrimmed Temporal Detection of Human-Object Interactions}
The proposed untrimmed temporal detection of human-object interactions task is related to previous research on untrimmed action detection. Existing approaches focus on one-stage methods, performing both temporal action detection and classification within a single network, aiming to identify actions without using action proposals. Recent works achieved state-of-the-art results using Vision Transformers. The authors of~\cite{zhang2022actionformer} proposed ActionFormer, a transformer network designed for temporal action localization in videos. This method estimates action boundaries through a combination of multiscale feature representation and local self-attention, which effectively models temporal dependencies.  
TriDet~\cite{shi2023tridet} uses a Trident-head to model the action boundary by estimating the relative probability distribution around the boundary. Features are extracted through a feature pyramid and aggregated with the proposed scalable granularity perception layer.

Other methods focused on masked video modeling for pretraining one-stage methods. In particular, InternVideo~\cite{wang2022internvideo} uses a combination of generative and discriminative self-supervised learning techniques by implementing masked video modeling and video-language contrastive learning. Recently, the authors of~\cite{wang2023videomae} proposed VideoMAE V2, which scaled VideoMAE~\cite{tong2022videomae} for building video foundation models through a dual masking strategy.

In this work, we assess the performance of state-of-the-art temporal action detection methods on the proposed ENIGMA-51 dataset considering ActionFormer~\cite{zhang2022actionformer}.

\subsection{Egocentric HOIs Detection} 
Several works have explored different aspects of human-object interactions (HOIs) from the egocentric perspective. The authors of~\cite{Hands_in_contact_Shan20} proposed a method based on the Faster-RCNN~\cite{ross2015faster} object detector to detect the hands and the objects present in the image, categorizing objects as either \textit{active} or \textit{passive}, determining the side of the hands (\textit{left} or \textit{right}), and predicting the contact state between the hand and the associated active object.
 
The authors of~\cite{ragusa2020meccano, ragusa2023meccano} investigated human-object interactions predicting bounding boxes around the active objects and the verb which describes the interaction exploiting multimodal signals with different instances of SlowFast networks~\cite{feichtenhofer2018slowfast}. The authors of~\cite{BenaventHOI2022} presented an architecture for detecting human-object interactions using two YOLOv4 object detectors~\cite{bochkovskiy2020yolov4} and an attention-based technique. The authors of~\cite{Grauman2022Ego4DAT} explored object transformations introducing the novel task of object state change detection and classification. While most of the analysis of human-object interactions relies on bounding box annotations, some works exploited hand poses and semantic segmentation masks~\cite{lu2021egocentric, VISOR2022}, contact boundaries~\cite{zhang2022fine}, which represents the spatial area where the interaction occurs, or additional modalities, such as depth maps and instance segmentation masks, to learn more robust representations~\cite{leonardi2023exploiting}.

In this work, we evaluate the HOIs detection method proposed in~\cite{leonardi2023exploiting} exploiting the fine-grained human-object interaction annotations of the ENIGMA-51.

\subsection{Short-Term Object Interaction Anticipation}

Past works addressed different variants of the short-term object interaction anticipation task. The authors of~\cite{Furnari2017NextactiveobjectPF} focused their study on the prediction of the next-active objects by analyzing their trajectories over time. 
The authors of~\cite{JIANG2021212} proposed a model that exploits a predicted visual attention probability map and the hands’ positions to predict next-active objects. The authors of~\cite{Dessalene_forecasting_contact} predicted future actions exploiting hand-objects contact representations. In particular, the proposed approach predicts future contact maps and segmentation masks, which are exploited by the Egocentric Object Manipulation Graphs framework~\cite{Dessalene2020EgocentricOM} for predicting future actions. 
The short-term object interaction anticipation task has been more formally defined in~\cite{Grauman2022Ego4DAT}. To tackle the task, the authors of~\cite{Grauman2022Ego4DAT} released a two-branch baseline composed of an object detector~\cite{ross2015faster} to detect next-active objects and a SlowFast~\cite{feichtenhofer2018slowfast} 3D network to predict the verb and the time to contact. The proposed baseline was extended by the authors of~\cite{Chen2022InternVideoEgo4DAP} who replaced Faster-RCNN with DINO~\cite{Zhang2022DINODW}, and SlowFast with a VideoMAE-pretrained transformer network~\cite{tong2022videomae}. Recently, StillFast an end-to-end approach has been proposed by~\cite{ragusa2023stillfast}. The method simultaneously processes a high-resolution still image and a video with a low spatial resolution, but a high temporal resolution. Recent state-of-the-art performances have been achieved by~\cite{pasca2023summarize} exploiting language. They proposed a multimodal transformer-based architecture able to summarise the action context leveraging pre-trained image captioning and vision-language models.

Due to its end-to-end training ability, in this work, we used StillFast~\cite{ragusa2023stillfast} as a baseline for the short-term object interaction anticipation benchmark on ENIGMA-51.

\subsection{Natural Language Understanding of Intents and Entities} 
Understanding intents and entities from text to extract knowledge about human-object interactions in the industrial domain is a task that has not been explored due to the lack of public egocentric datasets suitable for this task.

The authors of~\cite{vanzo2019hierarchical} addressed both intent classification and slot filling as a seq2seq problem, using an architecture that takes text input, generates ELMo embeddings~\cite{sarzynska2021detecting}, and incorporates one BiLSTM~\cite{schuster1997bidirectional} and self-attention layers for each task, outputting task-specific BIO (Beginning Inside and Outside) labels. 
In~\cite{chen2019bert} the BERT architecture has been explored to tackle the limited generalization capabilities of natural language understanding and propose a joint intent and classification architecture. The authors of~\cite{bunk2020diet} incorporate pre-trained word embeddings from language models and combine them with sparse word and character level n-grams features alongside a Transformer architecture.

While some works use speech-to-text models to convert speech input into text, others handle speech directly (Spoken Language Understanding). Earlier approaches proposed RNN-based or LSTM-based contextual SLU~\cite{shi2015contextual, HakkaniTr2016MultiDomainJS} which take into account previously predicted intents and slots. The authors of~\cite{haihong2019novel} proposed a BiLSTM-based architecture to manage the interrelated connections between intent and slots. In~\cite{xu2023efficient} has been introduced the Token-and-Duration Transducer (TDT) architecture for Automatic Speech Recognition (ASR), able to jointly predict both a token and its duration, enabling the skipping of input frames during inference based on the predicted duration output, resulting in significantly improved efficiency.

Since the ENIGMA-51 dataset comprises textual instructions about the activities performed by subjects, we exploited this textual information to explore the task of predicting intents and entities to extract knowledge about human-object interactions in the industrial domain.
\section{The ENIGMA-51 Dataset}
In our ENIGMA laboratory there are 25 different objects that can be grouped into fixed objects (such as an \textit{electric panel}) and movable objects (such as a \textit{screwdriver}). Differently than other egocentric datasets~\cite{ragusa2023meccano, sener2022assembly101} that contain industrial-like objects without textures, ENIGMA-51 includes real industrial objects as shown in Figure~\ref{fig:labels}. The complete list of the objects present in the ENIGMA laboratory is reported in the supplementary material. 

\subsection{Data Acquisition}
\label{sec:data_acquisition}
To collect data suitable to study human behavior in industrial domain, we designed two procedures consisting of instructions that involve humans interacting with the objects present in the laboratory to achieve the goal of repairing two electrical boards (see Figure~\ref{fig:labels} for visual examples). In particular, we designed two repairing procedures, one for each electrical board (\textit{high and low voltage}), with the help of industrial experts. For each procedure, we considered 4 different versions varying the use of a \textit{screwdriver} or \textit{electric screwdriver} and the electrical component to solder (\textit{resistor, capacitor or transformer}). Each procedure is composed of more than 100 steps, referencing objects and actions that were expected to be carried out in the industrial laboratory such as \textit{Turn on the welder using the switch on the corresponding socket (second from right)} and \textit{Set the temperature of the welder to 480 °C using the yellow “UP” button}. Based on these instructions, we developed a custom Microsoft HoloLens 2~\cite{Holo2} application which provided the instructions through audio, images and AR during the acquisition phase\footnote{Additional information about the repairing procedures are available in the supplementary material.}.
Considering that we designed two different repair procedures, each subject acquired at least one repairing video for each electric board obtaining a total of 51 videos. The 19 participants had different levels of experience in repairing electrical boards and using industrial tools. An example of the captured data is reported in Figure~\ref{fig:labels}. For each participant, we acquired the RGB stream from the Microsoft HoloLens 2 with a resolution of 2272x1278 pixels with a framerate of \textit{30 fps}. 
The average duration of the captured videos is 26.32min for a total of 22 hours of videos. We also synchronized the audio instructions with the captured video by assigning a timestamp when the user moved to the next instruction. 

\begin{table}[t]
\centering
\resizebox{0.8\columnwidth }{!}{
    \begin{tabular}{lrrrr}
    \textbf{Splits}             & \textbf{Train}    & \textbf{Val}  & \textbf{Test} & \textbf{Total}    \\ \hline
    \# Videos                   & 27                & 8             & 16            & 51                \\
    \# Videos Length            & $\simeq$11h       & $\simeq$4h    & $\simeq$7h    & $\simeq$22h       \\
    \# Images                   & 25,311             & 8,528          & 11,666         & 45,505             \\ 
    \# Objects                  & 152,865            & 53,486         & 68,784         & 275,135            \\ 
    \# Active Objects           & 4,709              & 1,700          & 2,933          & 9,342              \\ 
    \# Hands                    & 31,249             & 11,322         & 13,902         & 56,473             \\ 
    \# Hands in contact         & 5,039              & 1,833          & 3,171          & 10,043             \\ 
    \# Interactions frames      & 6,386              & 2,150          & 4,061          & 12,597             \\ 
    \# Interactions             & 7,133              & 2,406          & 4,497          & 14,036             \\ 
    \# Past frames              & 19,090             & 6,437          & 7,683          & 33,210             \\ 
    \# Next Object Interactions & 21,535             & 7,280          & 8,499          & 37,314             \\ 
    \hline
    \end{tabular}
}
\caption{Statistics of the ENIGMA-51 dataset considering the Training, Validation and Test splits.}
\label{tab:dataset_statistics}
\vspace{-6mm}
\end{table}

\subsection{Data Annotation}
\label{sec:data_annotation}
\begin{figure*}[ht]
    \centering
    \begin{subfigure}{0.48\textwidth}
        \centering
        \includegraphics[width=\linewidth]{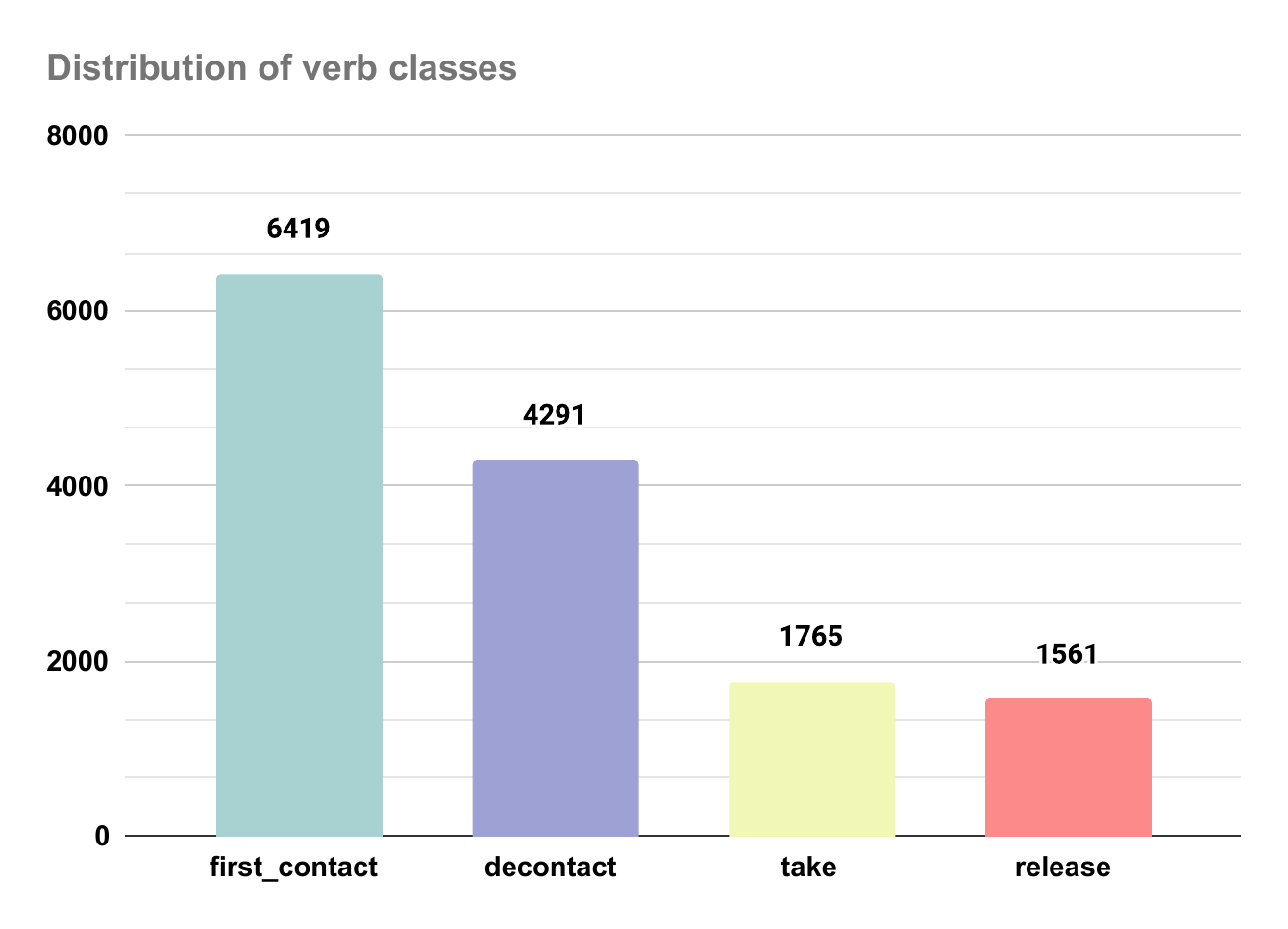}
    \end{subfigure}%
    \begin{subfigure}{0.48\textwidth}
        \centering
        \includegraphics[width=\linewidth]{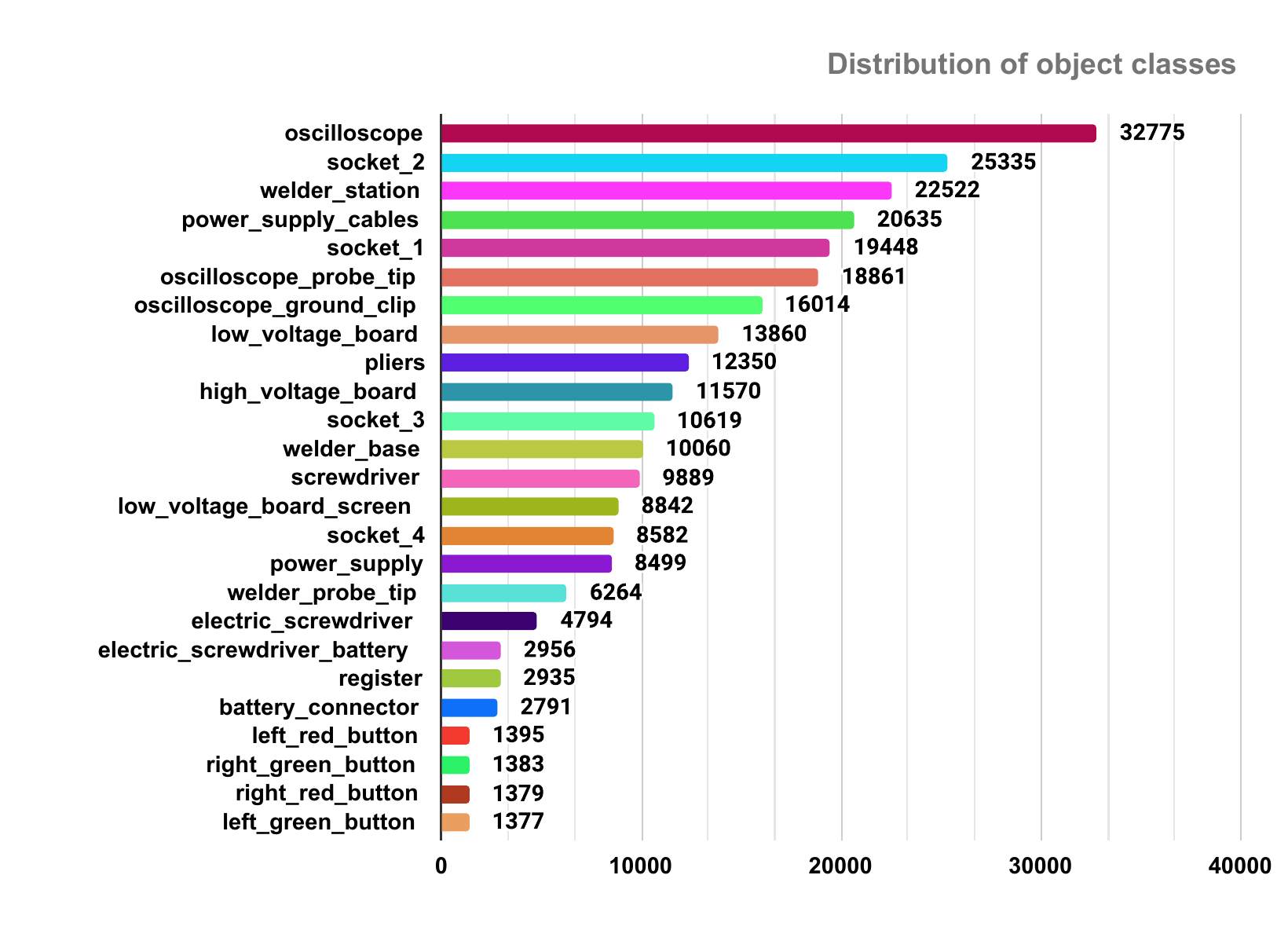}
    \end{subfigure}
    \caption{Distribution of verb (left) and object (right) classes over the 51 videos composing the ENIGMA-51 dataset.}
    \label{fig:stats}
\end{figure*}

We labelled the ENIGMA-51 dataset with a rich set of fine-grained annotations that can be used and combined to study different aspects of human behavior. Table~\ref{tab:dataset_statistics} summarizes statistics about the collected dataset.
\\
\textbf{Temporal and Verb Annotations:}
We identified all \textit{interaction key frames} in the 51 videos. For each identified interaction key frame, we assigned a timestamp and a verb describing the interaction. Our verb taxonomy is composed of 4 verbs: \textit{first-contact, de-contact, take, and release}. The 4 considered verbs represent the basic actions that a user performs to interact with objects. Note that the difference between \textit{first-contact} and \textit{take} is that \textit{first-contact} happens when the hand touches an object without taking it (e.g., pressing a button), while \textit{de-contact} is the first frame in which the hand-object contact breaks (e.g., end of pressing a button) and \textit{release} when the object is no longer held in the hand (e.g., put the screwdriver on the table). With this procedure, we annotated 14,036 interactions. Figure~\ref{fig:labels} reports an example of an interaction key frame with all the provided annotations, while Figure~\ref{fig:stats}-left shows the verbs distribution in the 51 videos.
\\
\textbf{Object Annotations:}
We considered 25 object classes which include both fixed (e.g., electric panel, oscilloscope) and movable objects (e.g., screwdriver, pliers) to assign a class to the objects present in the interaction key frames and in the past frames\footnote{Additional details about our object taxonomy are available in the supplementary material}. Each object annotation consists in a tuple $(class, x, y, w, h, state)$, where $class$ represents the class of the object, $(x, y, w, h)$ are the
2D coordinates which define the bounding box around the object in the frame, and the $state$ indicates if the object is involved in an interaction or not (\textit{active object vs. passive object}). With this annotation procedure, we annotated 275,135 objects. Figure~\ref{fig:stats}-right reports the distributions of the objects over the 51 videos of the ENIGMA-51 dataset. 
\\
\textbf{Hands Annotations:}
We annotated hands bounding boxes in the interaction key frames and in past frames. To speed up this annotation process, we generated pseudo-labels by processing the interaction key frames with a hand-object detector~\cite{Hands_in_contact_Shan20}, considering only the information related to the hands. Then, the annotators manually refined the bounding boxes, correcting the side of the hand and associating the hand with the previously labelled active object. Following this procedure, we labelled a total of 56,473 hands.
\\
\textbf{EHOI Annotations:}
For each of the interaction key frames, we considered: 1) hands and active object bounding boxes, 2) hand side (left and right), 3) hand contact state (contact and no contact), 4) hand-object relationships, and 5) object categories. For each hand, we assigned the \textit{hand contact state} to \textit{contact} if the hand was involved in an interaction of the type \textit{first-contact} or \textit{take}, and \textit{no-contact} for the \textit{release} and \textit{de-contact} categories. Additionally, to make the annotations consistent and uniform, we assigned the \textit{hand contact state} to \textit{contact} even for the hands that were already in physical contact with objects. Following this procedure, we annotated 12,597 interaction frames, 17,363 hands of which 10,043 were in contact, and 9.342 active objects.
\\
\textbf{Next Object Interaction Annotations:}
\label{sec:next_object}
Starting from the interaction key frame, we sampled frames every 0.4 seconds going backward up to 1.2 seconds before the beginning of the interaction timestamp. With this sampling strategy, we obtained 33210 past frames.  
We annotated the past frames with next object interaction annotations which consists of a tuple $(class,x,y,w,h,state,ttc)$ where $class$ represents the class of the object, $(x,y,w,h)$ are the 2D bounding box coordinates, $state$ indicates if the objects will be involved in an interaction and $ttc$ is a real number which indicates the time in seconds between the current timestamp and the beginning of the interaction. Figure~\ref{fig:labels} - bottom-left shows an example of labelled past frames.
\\
\textbf{Utterances:} 
Based on the instructions used for the acquisition of the dataset, we collected 265 textual utterances, which represent the types of questions that a worker might pose to a supervisor colleague while following a procedure within an industrial setting such as \textit{``How can I use the oscilloscope?''} or \textit{``Which is the next step that I do?''}. We manually annotated user goals as ``intents'' (e.g. ``object-instructions'') and key information as ``entities'' (e.g. ``object'') considering 24 intent classes and 4 entity types\footnote{\label{utterances}See the supplementary material for more details.}.
To enrich this set of utterances, we generated similar synthetic data by interacting with ChatGPT~\cite{chatGPT}. This study resulted in the creation of 100 unique utterances for each intent\footref{utterances}. The generated data was divided into three sets, G10, G50, and G100 which contain respectively 10, 50, and 100 generated unique utterances for each intent. Note that, all the utterances in G10 are also in G50 and G100, and all the utterances in G50 are also in G100.
\\
\textbf{Additional Resources:}
In order to enrich the ENIGMA-51 dataset, we release a set of resources useful to improve the impact of the dataset. We provide segmentation masks for the hands and the objects using SAM-HQ~\cite{sam_hq} and the 2D pose for the hands with MMPOSE~\cite{mmpose2020}. We also extracted visual representations through DINOv2~\cite{oquab2023dinov2} and CLIP~\cite{Radford2021LearningTV}. 
The 3D models of ENIGMA Laboratory and of all industrial objects within it have been acquired using the Matterport~\cite{Matterport} and ARTEC EVA~\cite{ARTEC} scanners, to enable the use of synthetic data to train scalable methods\footref{utterances}.

\section{Benchmark and baselines results}

\begin{table*}[t!]
    \centering
    \resizebox{0.8\textwidth}{!}{%
    \begin{tabular}{l*{10}{c}c}
              \textbf{Setting} & \multicolumn{10}{c}{\textbf{p-mAP (\%) temporal offset threshold (s)}} & \textbf{mp-mAP (\%)} \\ \hline
         & 1 & 2 & 3 & 4 & 5 & 6 & 7 & 8 & 9 & 10 &  \\ \hline
        “take vs. release” & 27.40 & 32.97 & 36.88 & 40.08 & 42.15 & 43.70 & 45.52 & 47.48 & 48.81 & 49.50 & 41.45 \\ 
        “first contact vs. de-contact” & 56.97 & 59.93 & 62.43 & 64.22 & 66.09 & 67.78 & 69.35 & 70.93 & 72.40 & 74.02 & 66.41 \\ 
        “all interactions” & 29.64 & 31.69 & 33.28 & 34.60 & 35.91 & 36.96 & 37.95 & 38.88 & 39.84 & 40.58 & 35.93 \\ \hline
    \end{tabular}
    }
    \caption{Comparisons of p-mAP under different temporal offset thresholds on 3 different interaction settings.}
    \label{tab:actionformer_results}
\end{table*}

\begin{table}[t]
    \centering
    \resizebox{\columnwidth}{!}{
    \begin{tabular}{ccccc} 
         \textbf{AP Hand}  & \textbf{AP H.+Side}   & \textbf{AP H.+State}  & \textbf{mAP H.+Obj}   & \textbf{mAP H.+All} \\ \hline
        90.81             & 90.35 & 73.31     & 46.51                 & 46.24 \\
        \hline
    \end{tabular}}
    \caption{Results of the baseline for the EHOI detection task.}
    \label{tab:ehoi_results}
\end{table}

\subsection{Untrimmed Temporal Detection of Human-Object Interactions}
\paragraph{\textbf{Task:}}
We consider the problem of detecting 4 basic human-object interactions (\textit{“take”}, \textit{“release”}, \textit{“first-contact”}, and \textit{“de-contact”}) from the untrimmed egocentric videos of the ENIGMA-51 dataset. Differently from the standard definition of untrimmed action detection, in this task, a prediction is represented as a tuple $(\hat c, \hat t_k, s)$, where $\hat c$ and $\hat t_k$ are respectively the predicted class and key timestamp (the timestamp of the interaction key frame) and $s$ is a confidence score.
\\
\textbf{Evaluation Measures:}
We evaluated our baselines using point-level detection mAP (p-mAP)~\cite{shou2018online}. We considered predictions as correct when they satisfied two criteria: 1) the interaction class matched the ground truth and 2) the difference between the predicted and ground truth timestamps is within a certain temporal threshold. We considered different temporal offset thresholds ranging from 1 to 10 seconds with an increment of one second~\cite{gao2019startnet,gao2021woad}; we averaged these values obtaining the mp-mAP values. 
\\
\textbf{Baseline:}
Our baseline for this task is based on ActionFormer~\cite{zhang2022actionformer}.  
It takes the pre-extracted video features as input and gives action boundaries (start and end timestamps) as outputs. Given our focus on predicting the timestamp when the HOI occurs, we considered only the predicted action start\footnote{\label{untrimmed}Additional information on implementation details, experiments, and results are reported in the supplementary material.} as output given by ActionFormer.
\\
\textbf{Results:}
Table~\ref{tab:actionformer_results} reports the results of the baseline.
We considered three variants of the task: 1) detecting only \textit{contact} and \textit{de-contact} interactions (first row), 2) considering only \textit{take} and \textit{release} interactions (second row), and 3) considering all the four interactions (third row). 
We obtained mp-mAP values of 41.45\%, 66.41\%, and 35.93\%, respectively, for \textit{“take vs. release”}, \textit{“first contact vs. de-contact”}, and \textit{“all interactions”}. The results highlight that detecting \textit{“take”} and \textit{“release”} interactions (first row) are more challenging compared to finding \textit{“first contact”} and \textit{“de-contact”} interactions (41.45\% vs. 66.41\%) due to the different semantic complexity. Moreover, when all the four interactions are considered, the performance decreases, obtaining a mp-mAP of 35.93\%\footref{untrimmed}.

\subsection{Egocentric HOI Detection}
\paragraph{\textbf{Task:}}
We consider the problem of detecting EHOIs from egocentric RGB images following the task definition proposed in~\cite{Hands_in_contact_Shan20,leonardi2023exploiting}. Given an input image, the aim is to predict the triplet \textit{\textless hand, hand contact state, active object\textgreater}. Additional details about the task are reported in~\cite{Hands_in_contact_Shan20,leonardi2023exploiting}. 
\\
\textbf{Baselines:}
The adopted baseline is based on the method proposed in~\cite{Hands_in_contact_Shan20}. We used the implementation proposed in~\cite{leonardi2023exploiting} which extends a two-stage object detector with additional modules that exploit hand features to predict the \textit{hand contact state} (in contact or not in contact), the \textit{side of hand} (left and right), and an \textit{offset vector} that indicates which object the hand is interacting with. Since the considered baseline is able to detect at most one contact per hand, we selected a subset of the $12,597$ interaction frames. This subset contains $15,955$ hands of which $8,753$ are in contact with an object, for a total of $7,680$ active objects.
\\
\textbf{Evaluation Measures:}
We used the following metrics based on standard \textit{Average Precision}~\cite{Hands_in_contact_Shan20, leonardi2023exploiting}: 1) \textit{AP Hand}: \textit{AP} of the hand detections, 2) \textit{AP Hand+Side}: \textit{AP} of the hand detections considering the correctness of the hand side, 3) \textit{AP Hand+State}: \textit{AP} of the hand detections considering the correctness of the hand state, 4) \textit{mAP Hand+Obj}: \textit{mAP} of the \textit{\textless hand, active object\textgreater} detected pairs, and 5) \textit{mAP Hand+All}: combinations of \textit{AP Hand+Side}, \textit{AP Hand+State}, and \textit{mAP Hand+Obj} metrics.
\\
\textbf{Results:}
Table~\ref{tab:ehoi_results} reports the results obtained with the proposed baseline.
Results show that the baseline achieved a \textit{AP Hand} of $90.81\%$, a \textit{AP Hand + Side} of $90.35\%$ ), a \textit{mAP H.+State} of $73.31\%$, a \textit{mAP H.+Obj} of $46.51\%$ and a \textit{mAP H.+All} of $46.24\%$, pointing out that the use of domain-specific data in training is needed to exploit the knowledge of the industrial objects to support workers in the industrial domain.


\subsection{Short-Term Object Interaction Anticipation} 
\paragraph{\textbf{Task:}}
The short-term object interaction anticipation task~\cite{Grauman2022Ego4DAT} aims to detect and localize the next-active objects, to predict the verb that describes the future interaction, and to determine when the interaction will start. Formally, the task consists in predicting future object interactions from a video $V$ and a timestamp $t$. The models can only use the video frames up to time $t$ and have to produce a set of predictions for the object interactions that will occur after a time interval $\delta$. Predictions consist of a bounding box over the next-active objects, a noun label, a verb label describing the future interaction, a real number indicating how soon the next interaction will start, and a confidence score. 
\\
\textbf{Evaluation Measures:}
We evaluated the model’s performance with Top 5 mean Average precision measures~\cite{Grauman2022Ego4DAT} that capture different aspects of the task: Top-5 mAP Noun, Top-5 mAP Noun+Verb, Top-5 mAP Noun+TTC, and Top-5 mAP Noun+Verb+TTC, which is also referred to as Top-5 mAP Overall.
\\
\textbf{Baseline:}
We adopted StillFast~\cite{ragusa2023stillfast} as the baseline\footnote{\url{https://github.com/fpv-iplab/stillfast}}. The model has been designed to extract 2D features from the considered past frame and 3D features from the input video clip. Feature stacks are merged through a combined feature pyramid layer and sent to the prediction head which is based on the Faster-RCNN head~\cite{ross2015faster}. 
Features are fused and used to predict object (noun), verb probability distributions and time-to-contact (ttc) through linear layers along with the related prediction score \textit{s}.
\\
\textbf{Results:}
Table~\ref{tab:res_STA} reports the results on test set of the ENIGMA-51 dataset considering the Top-5 mAP measures. StillFast obtains a Noun Top-5 mAP of 78.79\%, demonstrating the ability to detect and classify the next-active objects processing images and videos simultaneously. When verbs and time to contact are predicted, performance drops according to Noun+Verb Top-5 mAP of 62.58\%, Noun+TTC Top-5 mAP of 35.77\%, and Overall Top-5 mAP of 27.83\%. Qualitative results are reported in the supplementary material.

\begin{table}[t]
\centering
\resizebox{0.7\columnwidth}{!}{
\begin{tabular}{cccc}
\textbf{Noun} & \textbf{N+V} & \textbf{N+TTC} & \textbf{Overall} \\ \hline
78.79 & 62.58 & 35.77 & 27.83 \\ \hline
\end{tabular}
}
\caption{Results\% in Top-5 mean Average Precision for the Short-Term Object Interaction Anticipation task. N stands for noun, N+V stands for Noun+Verb and N+TTC stands for Noun+Time to Contact.}
\label{tab:res_STA}
\vspace{-6mm}
\end{table}

\subsection{NL Understanding of Intents and Entities} 
\paragraph{\textbf{Task:}}
We considered the problem of classifying the intent of a user utterance, falling into one of the considered 24 classes, as well as the problem of entity slot filling, including four different slot types: \textit{``object''}, \textit{``board''}, \textit{``component''} and \textit{``procedure''}.
Given an input utterance \textit{U}, the task is to predict the intent class \textit{i}, and to detect any entities \textit{e}, if present, as well as the slot types \textit{t} associated to them, outputting zero or more \textit{$<$e, t$>$} couples. The complete list of intents/entities is reported in the supplementary material.
\\
\textbf{Evaluation Measures:}
We evaluate the baseline using the standard accuracy, and F1-score evaluation measures. 
\\
\textbf{Baseline:}
The baseline is based on the DIETClassifier~\cite{bunk2020diet}. We performed the tokenization and featurization steps before passing the utterances to the model. Specifically, we used the SpacyNLP, SpacyTokenizer, CountVectorsFeaturizer, SpacyFeaturizer and DIETClassifier components offered by the Rasa framework~\cite{rasa}. 
\\
\textbf{Results:}
Table~\ref{tab:intententityresults} reports the results obtained for intent and entity classification. Five different variants of the training set (see Section~\ref{sec:data_annotation}) were explored: real data, real data + G10 data, real data + G50 data, real data + G100 data, and G100.
The best results for the intent classification have been obtained using only real data obtaining an accuracy of 0.867 and an F1-score of 0.844. The baseline suffers when generated data are included, which introduces noise and makes performance worse, reaching an accuracy of 0.584 (-0.283) and an F1-score of 0.564 (-0.280). These results suggest that, in this challenging industrial scenario, generative models, such as GPT~\cite{chatGPT} are not yet capable of generating appropriate data with regard to understand human's intent in this domain, and the use of manually annotated data is still necessary.  
Instead, considering the ability to predict the entities of human's utterances which represent more simple concepts with respect to the human's intents, only generated data (last row) are enough. In particular, the model trained with the G100 set obtains better performance than one trained only with real data (1.00 vs. 0.994 for accuracy and 1.00 vs. 0.981 for F1-score)\footnote{Additional details about the used prompting, the implementation details, and the results are reported in the supplementary material.}.

\begin{table}[t]
\centering
\resizebox{0.8\columnwidth}{!}{%
\begin{tabular}{ccccc}
\textbf{} & \multicolumn{2}{c}{\textbf{Intent}} & \multicolumn{2}{c}{\textbf{Entity}} \\ \cline{1-5} 
\textbf{Training} & \multicolumn{1}{l}{\textbf{Accuracy}} & \textbf{F1-score} & \multicolumn{1}{l}{\textbf{Accuracy}} & \textbf{F1-score} \\ \hline
real & \textbf{0.867} & \textbf{0.844} & 0.994 & 0.981 \\
real+G10 & 0.830 & 0.815 & 1.00 & 1.00 \\
real+G50 & 0.792 & 0.773 & 1.00 & 1.00 \\
real+G100 & 0.792 & 0.784 & 1.00 & 1.00 \\
G100 & 0.584 & 0.564 & \textbf{1.00} & \textbf{1.00} \\ \hline
\end{tabular}
}
\caption{Results for intents and entities classification considering different sets of training data.}
\label{tab:intententityresults}
\end{table}

\section{Conclusion}
We proposed ENIGMA-51, a new egocentric dataset acquired in an industrial environment and densely annotated to study human behavior. In addition, we performed baseline experiments aimed to study different aspects of human behavior in the industrial domain addressing four tasks. Existing methods show promising results but are still far from reaching reasonable performance to build an intelligent assistant able to support workers in the industrial domain. This opens up opportunities for future in-depth investigations.

\section*{\uppercase{Acknowledgements}}
This research is supported by Next Vision\footnote{Next Vision: https://www.nextvisionlab.it/} s.r.l., by MISE - PON I\&C 2014-2020 - Progetto ENIGMA  - Prog n. F/190050/02/X44 – CUP: B61B19000520008, and by the project Future Artificial Intelligence Research (FAIR) – PNRR MUR Cod. PE0000013 - CUP: E63C22001940006.

\section*{\uppercase{Supplementary Material}}
\label{sec:supp_material}
This document is intended for the convenience of the reader and reports additional information about the collection and the annotations of the proposed dataset, as well as implementation details of the adopted baselines. This supplementary material is related to the following submission:\\

\begin{itemize}
    \item F. Ragusa, R. Leonardi, M. Mazzamuto, C. Bonanno, R. Scavo, A. Furnari, G. M. Farinella. ENIGMA-51: Towards a Fine-Grained Understanding of Human-Object
Interactions in Industrial Scenarios. In IEEE Winter Conference on Applications of Computer Vision (WACV), 2024.
\end{itemize}
The reader is referred to the manuscript and to our web page \url{https://iplab.dmi.unict.it/ENIGMA-51/} to download the dataset and for further information.

\section{The ENIGMA-51 Dataset}
\section{The ENIGMA-51 Dataset}
\begin{figure*}[ht]
  \centering
    \includegraphics[width=0.85\linewidth]{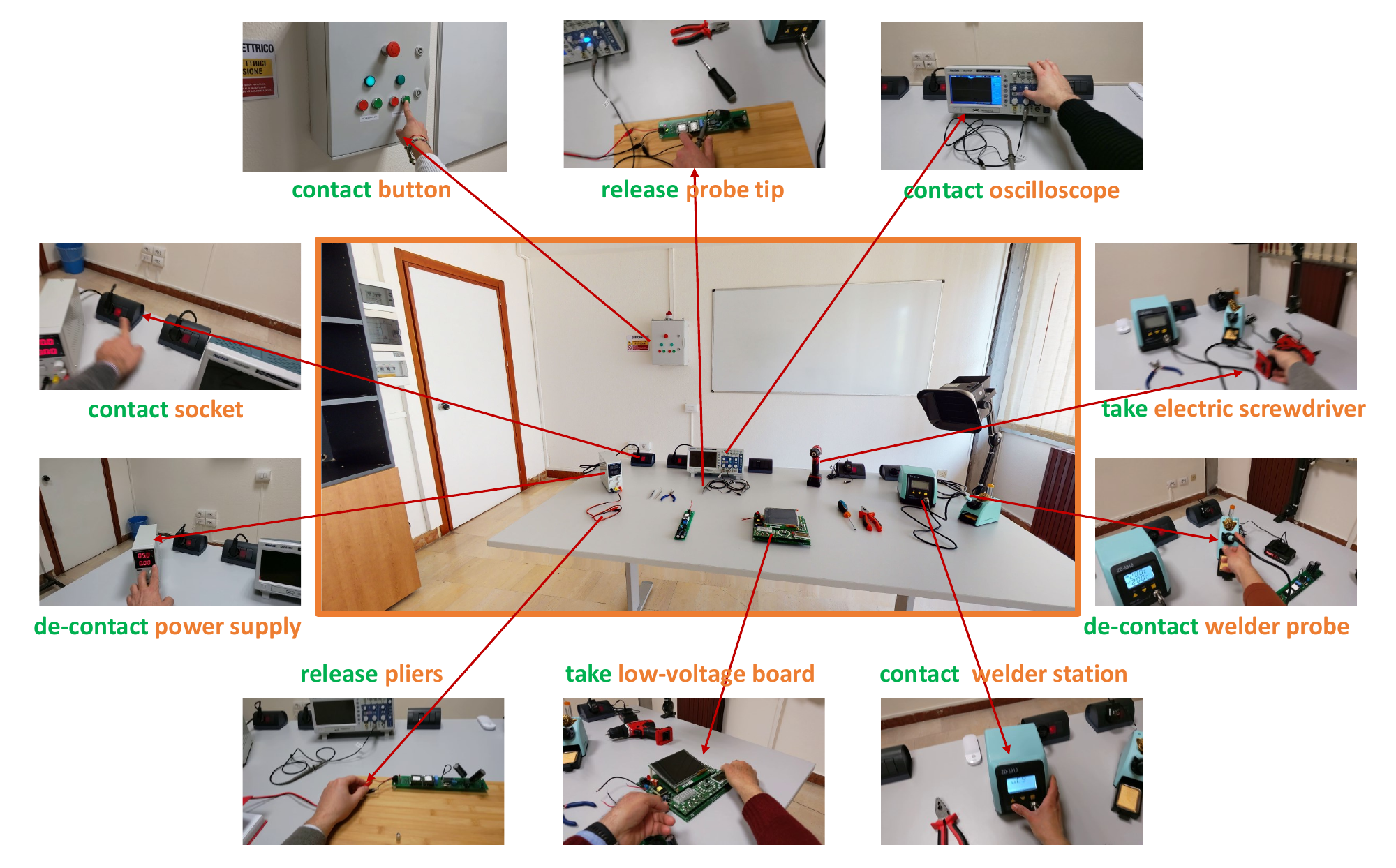}
    \caption{The ENIGMA-51 dataset has been acquired in an industrial laboratory. We show some interaction key frames with the related verb (in green) and the object involved in the interaction (in orange).}
    \label{fig:dataset}
\end{figure*}

The ENIGMA-51 dataset has been acquired in an industrial laboratory by 19 subjects who wore a Microsoft Hololens 2 providing audio instructions to follow to complete a repair procedure of electrical boards. The dataset is composed of 51 egocentric videos. Each video includes a complete repair procedure of an electrical board where movable objects (see Section~\ref{sec:object_annotations}) were placed in random positions on the working table. Each subject acquired at least one video for each electrical board (\textit{high} and \textit{low} voltage) obtaining a total of 51 videos. The dataset was divided into training, validation, and test sets. Each set contains videos acquired from different subjects, and there is no overlap between the subjects in any of the sets. Figure~\ref{fig:dataset} shows the industrial laboratory in which the ENIGMA-51 dataset has been acquired.

\subsection{Instruction for repair procedures}
We designed two procedures composed of instructions that involve humans interacting with the objects of the laboratory to achieve the goal of repairing two different electrical boards. These procedures have been designed with the support of industrial experts with the aim of capturing realistic human-object interactions in a real industrial domain. Specifically, we designed a procedure for each electrical board: \textit{High Voltage Repair} and \textit{Low Voltage Repair}. Then, for each procedure we forced the use of one of the electric or standard screwdrivers and the soldering of one of the resistor/capacitor/transformer electrical components, obtaining four variants of each procedure. Each procedure is designed to allow the worker to interact with all the industrial objects and electric machinery present in the laboratory. With the provided instructions, we expected users to interact with movable objects (e.g., \textit{``Take the soldering iron’s probe''} or \textit{``Place the electric board on the working area''}) and with fixed machinery (e.g., `\textit{`Press the two green buttons on panel A''} or \textit{``Adjust the voltage knob of the power supply to set a voltage of 5 Volts''}).
As example, Table~\ref{tab:procedure} reports a complete repair procedure for the low-voltage electric board using the standard screwdriver and soldering the capacitor.
To provide instructions to the user without the use of physical manuals, we developed an application for the Microsoft HoloLens 2.

\begin{table*}[ht!]
\centering
\resizebox{1\linewidth}{!}{%
\begin{tabular}{|c|p{0.45\linewidth}|c|p{0.45\linewidth}|c|p{0.45\linewidth}|}
\hline
\textbf{Step} & \textbf{Description} & \textbf{Step} & \textbf{Description} & \textbf{Step} & \textbf{Description} \\
\hline
1 & Sit at the workbench & 41 & Place the pliers on the workbench & 81 & Place the pliers on the workbench \\
2 & Pronounce the voice command ``Record" to start recording, and wait for the acoustic signal for confirmation & 42 & Lower the soldering iron temperature to the minimum (160°C) using the yellow ``DOWN" button & 82 & Place the pliers on the workbench \\
3 & Turn the lamp located on the workbench on and off while looking at it & 43 & Turn off the soldering iron using the socket switch & 83 & Turn off the soldering iron using the socket switch \\
4 & Exit the laboratory & 44 & Fix the board to the workbench using the screwdriver & 84 & Connect the display to the board \\
5 & Enter the laboratory and close the door & 45 & Observe the power supply & 85 & Turn on the power supply using the socket switch \\
6 & Go to panel A and observe it for a moment & 46 & Adjust the current knob of the power supply until the green LED lights up & 86 & Adjust the power supply voltage knob to set a voltage of 5 Volts \\
7 & Press the two green buttons on panel A & 47 & Connect the power supply cables to the board's power points & 87 & Connect the power supply cables to the board's power points \\
8 & Head back to the workbench and sit down & 48 & Observe the board for a few seconds to verify the red LED turning on & 88 & Observe the power supply \\
9 & Observe the low-voltage board for a while & 49 & Turn off the power supply using the socket switch & 89 & Turn off the power supply using the socket switch \\
10 & Take the low-voltage board and place it on the work area & 50 & Set the current and voltage knobs of the power supply to 0 & 90 & Set the current and voltage knobs of the power supply to 0 \\
11 & Remove the screws from the workbench using the screwdriver & 51 & Disconnect the clip and probe of the oscilloscope & 91 & Disconnect the power supply cables \\
12 & Secure the board to the workbench using the screwdriver & 52 & Turn on the oscilloscope using the socket switch & 92 & Observe the board for a few seconds to verify the red LED turning on \\
13 & Observe the power supply & 53 & Activate channel 2 of the oscilloscope using the ``CH2 MENU" button with a blue outline & 93 & Turn on the oscilloscope using the socket switch \\
14 & Turn on the power supply using the socket switch & 54 & Connect the ground clip of the probe to test point 1 & 94 & Activate channel 2 of the oscilloscope using the ``CH2 MENU" button \\
15 & Adjust the current knob of the power supply until the green LED lights up & 55 & Use the probe tip to check for signals at test point 2 on the board & 95 & Connect the ground clip of the probe to test point 1 \\
16 & Adjust the voltage knob of the power supply to set a voltage of 5 Volts & 56 & Press the ``Auto Set" button on the oscilloscope & 96 & Use the probe tip to check for signals at test point 2 on the board \\
17 & Connect the power supply cables to the board's power points & 57 & Observe the oscilloscope's display & 97 & Press the ``Auto Set" button on the oscilloscope \\
18 & Observe the board for a few seconds to verify the red LED turning on & 58 & Rotate the ``position" knob above the ``CH2 MENU" button with a blue outline randomly & 98 & Observe the oscilloscope's display \\
19 & Turn off the power supply using the socket switch & 59 & Repeat the previous three steps for the remaining test points (from number 3 to number 7) & 99 & Rotate the ``position" knob above the ``CH2 MENU" button with a blue outline randomly \\
20 & Set the current and voltage knobs of the power supply to 0 & 60 & Set the current and voltage knobs of the power supply to 0 & 100 & Repeat the previous three steps for remaining test points (from number 3 to number 7) \\
21 & Disconnect the power supply cables & 61 & Disconnect the ground clip and probe from the oscilloscope & 101 & Turn off the power supply using the socket switch \\
22 & Remove the board from the workbench using the screwdriver & 62 & Deactivate channel 2 of the oscilloscope & 102 & Set the current and voltage knobs of the power supply to 0 \\
23 & Unscrew the 4 screws on the back of the board using the screwdriver & 63 & Turn off the oscilloscope using the socket switch & 103 & Disconnect the ground clip and probe from the oscilloscope \\
24 & Remove the display from the board & 64 & Remove the board from the workbench using the screwdriver & 104 & Deactivate channel 2 of the oscilloscope \\
25 & Observe the soldering iron & 65 & Observe the soldering iron & 105 & Turn off the oscilloscope using the socket switch \\
26 & Turn on the soldering iron using the socket switch & 66 & Turn on the soldering iron using the socket switch & 106 & Remove the board from the workbench using the screwdriver \\
27 & Set the soldering iron temperature to 200 degrees using the yellow ``UP" button & 67 & Set the soldering iron temperature to 200 degrees using the yellow ``UP" button & 107 & Observe the soldering iron \\
28 & Grab the black capacitor on the board with pliers & 68 & Grab the black capacitor on the board with pliers & 108 & Turn on the soldering iron using the socket switch \\
29 & Take the soldering iron's probe & 69 & Take the soldering iron's probe & 109 & Set the soldering iron temperature to 200 degrees using the yellow ``UP" button \\
30 & Touch the first pin of the black capacitor with the soldering iron's probe for 5 seconds & 70 & Touch the first pin of the black capacitor with the soldering iron's probe for 5 seconds & 110 & Head to panel A \\
31 & Touch the second pin of the capacitor for 5 seconds with the soldering iron's probe & 71 & Touch the second pin of the capacitor for 5 seconds with the soldering iron's probe & 111 & Observe panel A for a moment \\
32 & Place the pliers on the workbench & 72 & Place the pliers on the workbench & 112 & Press the two red buttons on panel A \\
33 & Place the soldering iron's probe & 73 & Place the soldering iron's probe & 113 & Head to the door \\
34 & Place the board vertically & 74 & Place the board vertically & 114 & Exit the laboratory \\
35 & Grab the black capacitor on the board with pliers & 75 & Grab the black capacitor on the board with pliers & 115 & Enter the laboratory \\
36 & Take the soldering iron's probe & 76 & Take the soldering iron's probe & 116 & Pronounce the voice command ``Stop" to end the recording, hear an acoustic signal for confirmation \\
37 & Touch the first pin of the black capacitor on the back of the board for 5 seconds & 77 & Touch the first pin of the black capacitor on the back of the board for 5 seconds & & \\
38 & Touch the second pin of the capacitor on the back of the board for 5 seconds & 78 & Touch the second pin of the capacitor on the back of the board for 5 seconds & & \\
39 & Place the soldering iron's probe & 79 & Place the soldering iron's probe & & \\
40 & Place the board on the work area & 80 & Place the board on the work area & & \\
\hline
\end{tabular}
}
\caption{Low Voltage Board Repair Procedure (Standard Screwdriver Version)}
\label{tab:procedure}
\end{table*}

\begin{figure}[ht!]
    \centering
    \includegraphics[width=0.99\linewidth]{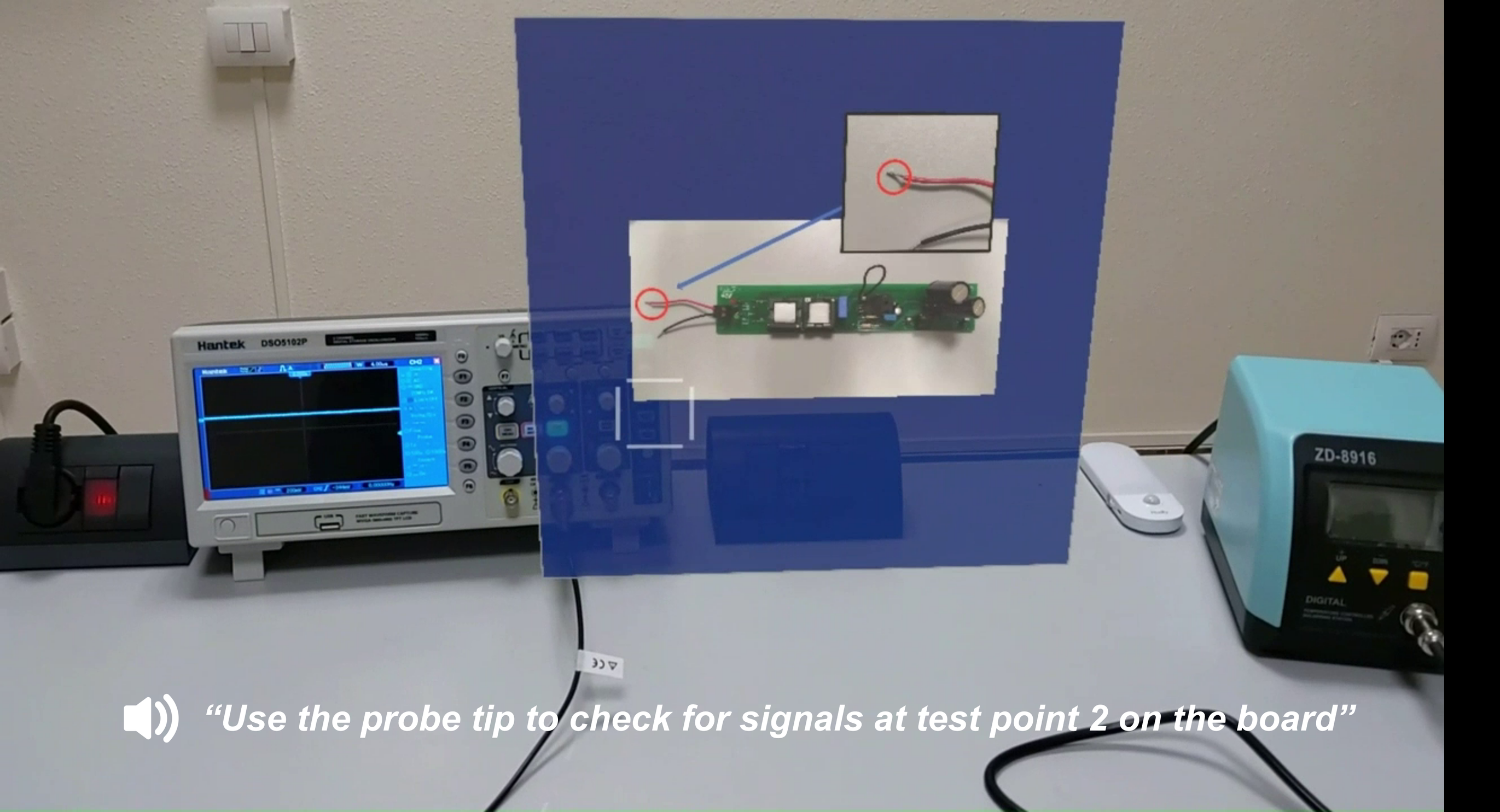}
    \caption{A screenshot captured from the developed application, during the acquisition phase.}
    \label{fig:hololens_app}
\end{figure}

\subsubsection{HoloLens2 acquisition application}
We developed an application for Microsoft HoloLens 2 using the Unit 3D graphic engine to provide instructions to the participants during the acquisition.
In particular, the application helps the operators during the acquisition phase, providing audio instructions and showing images to facilitate complex operations (e.g., where to connect the oscilloscope ground clip). Figure~\ref{fig:hololens_app} shows an example of the acquisition tool. The application integrates voice commands for the human-device interaction (e.g. the possibility to say ``next" or ``back" over the steps of the procedure). The audio guide describes the operations to be performed during the acquisitions. To create these audio tracks, a Python script has been created that utilizes the gTTS library for interacting with Google Translate APIs. This script takes an input text file divided into textual blocks and converts it into a set of MP3 audio tracks. After wearing the device, the operator will interact with the application using the following voice commands:
\begin{itemize}
    \item \textbf{Forward}: to play the audio track for the next set of instructions.
    \item \textbf{Backward}: to play the audio track of the previous set of instructions.
    \item \textbf{Repeat}: to replay the audio track for the current set of instructions.
    \item \textbf{Record}: Using this command, the operator starts video recording (the application will play a sound for confirmation).
    \item \textbf{Stop}: Using this command, the operator stops the video recording (the application will play a sound for confirmation).
\end{itemize}

\subsection{Data Annotation} \label{sec:dataset}
\subsubsection{Object Annotations} \label{sec:object_annotations}
\begin{figure*}[ht]
  \centering
    \includegraphics[width=.8\linewidth]{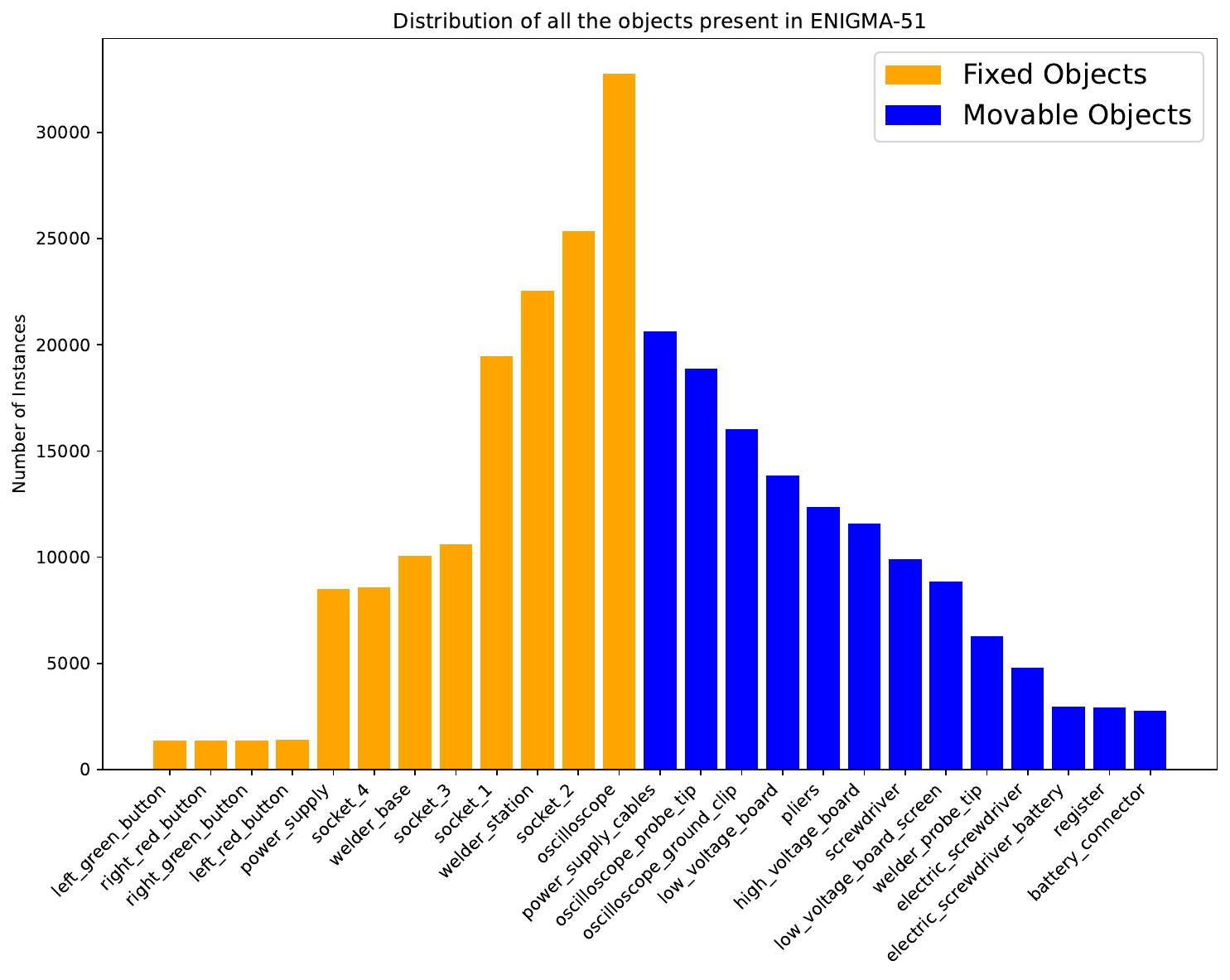}
  \caption{We report the object class distribution over the 51 videos of ENIGMA-51 grouping them into two categories: \textit{fixed} (orange) and \textit{movable} (blue).}  
   \label{fig:distribution}
\end{figure*}

\begin{figure}[ht!]
    \centering
    \includegraphics[width=0.99\linewidth]{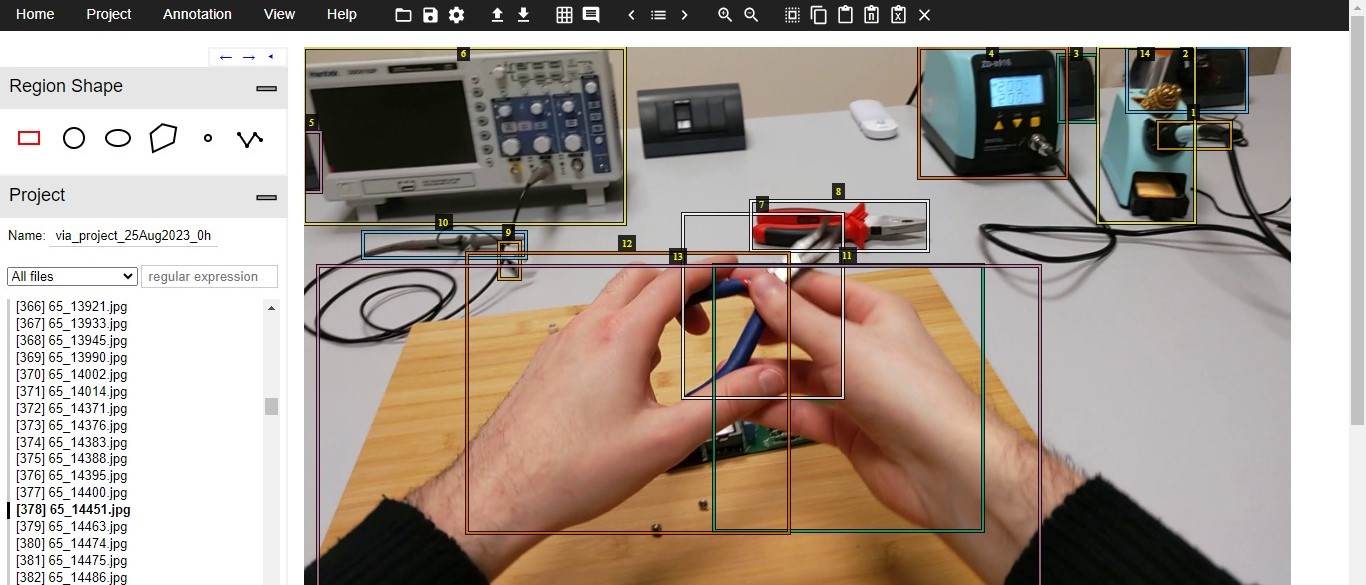}
    \caption{VGG Image Annotator tool.}
    \label{fig:annotation_tool}
\end{figure}

In our industrial setting, we considered both fixed (e.g., \textit{oscilloscope}, \textit{power suppl}) and movable objects (e.g., \textit{screwdriver}, \textit{electric boar}) present in the industrial laboratory. In particular, our object taxonomy is composed of 25 different objects: \textit{power supply, power supply cables, oscilloscope, oscilloscope probe tip, oscilloscope ground clip, welder station, welder base, welder probe tip, electric screwdriver, electric screwdriver battery, battery connector, screwdriver, pliers, high voltage board, low voltage board, low voltage board screen, register, left red button, left green button, right red button, right green button, socket 1, socket 2, socket 3, and socket 4}.
Figure~\ref{fig:distribution} reports the object class distribution grouping them into \textit{fixed} and \textit{movable} objects.
The dataset has been labelled manually by a group of annotators that used the VGG Image Annotator~\cite{VGG} with a custom project (see Figure\ref{fig:annotation_tool}).

\begin{table*}[t]
    \centering
    \begin{tabular}{ll}
       Intent & Description \\ \hline
       ``greet'' & Greet and start a conversation \\ 
        ``procedure-tutorial'' & Ask a specific question about the ongoing procedure \\
        ``object-warnings'' & Know if there are alerts for a specific object \\
        ``turn-object-on'' & Turn on an object \\
        ``turn-object-off'' & Turn off an object \\
        ``which-ppe-procedure'' & Know which PPE is required to perform a specific procedure \\
        ``which-ppe-object'' & Know which PPE is required to use a specific object \\
        ``object-instructions'' & Know how to use a specific object \\
        ``is-object-on'' & Find out if an object is turned on or off \\
        ``object-time'' & Find out how long an object has been used \\
        ``where-board'' & Know where a specific electronic board is located, or identify it on the working area \\
        ``board-detail'' & Know the location of a component on an electronic board \\ 
        ``where-object'' & Know where a specific object is located, or identify it on the working area \\
        ``object-detail'' & Know the location of a component on an object \\
        ``start'' & Start a procedure \\
        ``next'' & Hear the next step in the ongoing procedure \\
        ``previous'' & Hear the previous step in the ongoing procedure \\
        ``repeat'' & Hear the current step in the ongoing procedure \\
        ``all-objects'' & Know what objects are present in the laboratory \\
        ``ok-objects'' & Know what objects can be used \\
        ``on-objects'' & Know what objects are powered \\
        ``where-ppe'' & Know where the PPEs are located \\
        ``inform'' & Specify an entity \\
        ``out-of-scope'' & This category includes all questions that are not relevant to the previous intents \\ 
    \end{tabular}
     \caption{The 24 intent classes considered during our collection.}
    \label{tab:intents}
\end{table*}

\begin{table}[t]
    \centering
    \begin{tabular}{ll}
       Entity & Example \\ \hline
       ``object'' & [soldering iron](object) \\
       ``board' & [low voltage](board) \\ 
       ``component'' & [display](component) \\ 
       ``procedure'' & [repair](procedure) \\
    \end{tabular}
    \caption{The table reports our entity taxonomy composed of 4 classes.}
    \label{tab:entities}
\end{table}

\subsubsection{Utterances} 
Classifying intents and entities within the industrial domain can be beneficial in the development of intelligent assistants that support workers during their interactions and ensure enhanced workplace safety. 
Using the instructions that guided participants to acquire the ENIGMA-51 dataset, we obtained 265 textual utterances that simulate the kinds of questions a worker may have for a supervisor colleague as they carry out a procedure in an industrial setting. We manually labelled these utterances as ``intents'' (e.g. ``object-instructions'') considering a taxonomy of 24 classes and as ``entities'' (e.g. ``object'') considering 4 entity types. Table~\ref{tab:intents} reports the list of the 24 intent classes with an associated description. 
Each entity has been annotated using square brackets to denote its starting and ending characters in the text and round brackets to enclose the entity type. As a result, each entity is annotated following the [entity](type) form. Table~\ref{tab:entities} reports the list of the 4 considered entities.

To enrich this set of utterances, we generated similar utterances through the prompting of ChatGPT~\cite{chatGPT}, obtaining 100 unique utterances for each intent. Due to the unique structure of ``inform'' utterances, which consist of only an entity and optionally an article, generating a set of 100 utterances was unfeasible; hence, a total of 10 utterances for the ``inform'' intent were produced. 
The ``inform'' intent is defined for conversations in which a worker's question cannot be adequately answered solely by performing slot filling on their initial utterance. This is often the case when some of the required entities for formulating an appropriate response are missing. For example, in the following conversation: 
\begin{itemize}
    \item{Worker: \textit{What's this object? I don't know how to use it.}}
    \item{Assistant: \textit{Which object?}}
    \item{Worker: \textit{The oscilloscope.}}
\end{itemize}
The worker's first utterance falls under the ``object-instructions'' intent, whereas the worker's second utterance falls under the ``inform'' intent.
Examples of utterances belonging to the inform intent include \textit{``[high voltage](board) board [testing](procedure) procedure''} and \textit{``[screwdriver](object)''}.

The used prompt for each intent, except for ``inform'' and ``out-of-scope'' intents, was the following: \textit{``Imagine being an operator working inside an industrial laboratory. You can communicate with someone who knows the laboratory perfectly, including all the present objects and possible procedures that can be carried out. There are several intents you could have while operating within this industrial laboratory. This is one: $<$intent description$>$. Since you'll have to communicate with the other person through text messages, try to avoid all forms of greeting and politeness. For this intent, imagine 100 unique sentences you would say to your interlocutor to express your intent and achieve the desired result.''} Please note that $<$intent description$>$ was replaced with the description of each specific intent, using the descriptions listed in Table~\ref{tab:intents}. Exceptions were made for the ``inform'' intent, for which we prompted the model to generate 10 unique sentences, and the ``out-of-scope'' intent, for which we used the following prompt: \textit{``Imagine being an operator working inside an industrial laboratory. You can communicate with someone who knows the laboratory perfectly, including all the present objects and possible procedures that can be carried out. There are several intents you could have while operating within this industrial laboratory, which I will list below: $<$full list of intent descriptions$>$. Since you'll have to communicate with the other person through text messages, try to avoid all forms of greeting and politeness. Knowing these intents, generate 100 unique sentences that are out of scope.''} Please note that $<$full list of intent descriptions$>$ was replaced with the full list of intent descriptions listed in Table~\ref{tab:intents}. Examples of the obtained utterances include: \textit{``Provide [high voltage](board) board [repair](procedure) procedure tutorial now.''}, \textit{``Quick status check: alerts for [battery charger](object)?''}, \textit{``I require an image of the [display](component) that belongs to the [low voltage](board) board.''}, \textit{``Where's the PPE kept?''}, \textit{``[high voltage](board) board [testing](procedure) procedure''} and \textit{``I need help with my car's engine trouble; can you assist me?''} for the ``procedure-tutorial'' ``object-warnings'' ``board-detail'' ``where-PPE'', ``inform'', ``out-of-scope'' respectively. 

As ChatGPT was not able to generate 100 unique utterances, we carried out additional duplicate filtering and re-prompted the model in order to generate more utterances, until we met the criteria of gathering 100 unique utterances for each intent. We hypothesize that the inability to generate a set of unique utterances is due to the many constraints expressed in our prompt, which on the other hand was designed to generate utterances that reflected the real ones collected in the same laboratory setting.

\subsection{Additional Resources}
\subsubsection{3D models of the laboratory and objects}
\begin{figure}[t!]
    \centering
    \includegraphics[width=0.9\linewidth]{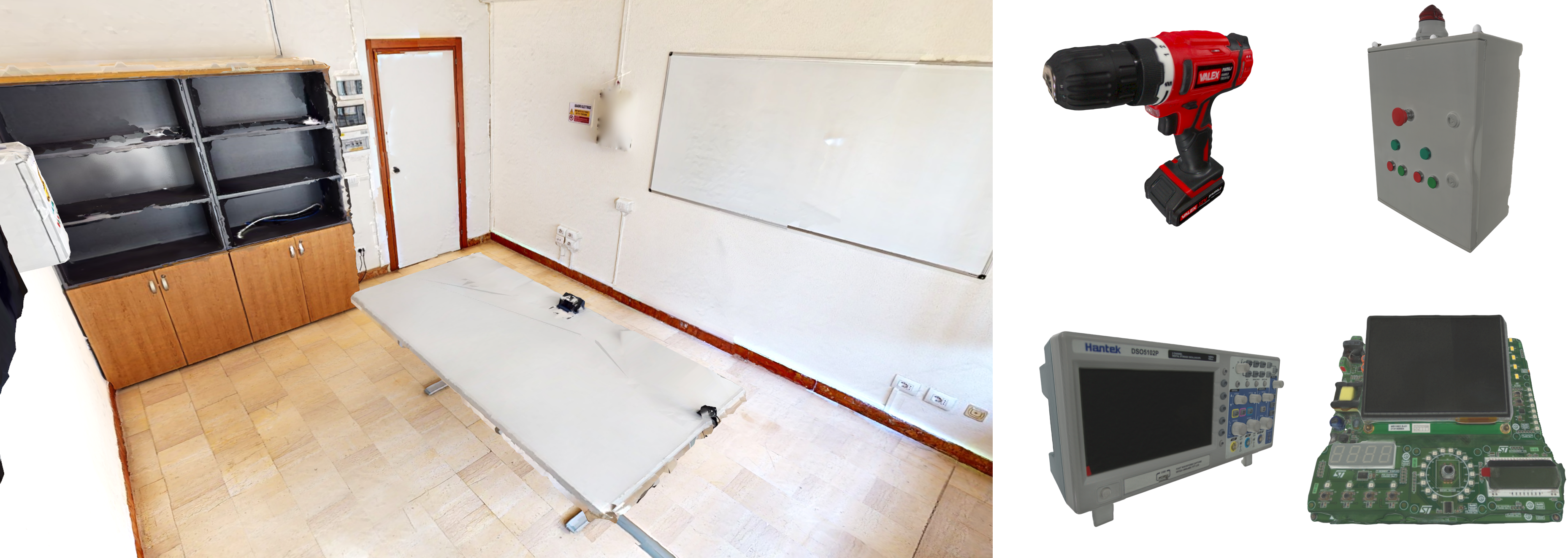}
    \caption{The acquired 3D models of the laboratory and some industrial objects within the set of ENIGMA-51.}
    \label{fig:lab_3d}
\end{figure}

To enable the use of synthetic data to train scalable methods, we acquired the 3D models of the laboratory and all the 25 industrial objects. We used two different 3D scanners to create 3D models. Specifically, we
used the structured-light 3D scanner Artec Eva\footnote{\url{https://www.artec3d.com/portable-3d-scanners/artec-eva}} for scanning
the objects, and the MatterPort\footnote{\url{https://matterport.com/}} device to scan the industrial laboratory.
Figure~\ref{fig:lab_3d} illustrates the 3D models of the laboratory and some industrial objects within the set of ENIGMA-51. The 3D model of the laboratory weighs 30MB and covers an area of approximately 20 square meters, instead, the weight of the object's 3D model varies from $5$ to $20$ MB. 

\subsubsection{Hands and Objects Segmentation using SAM-HQ}
\begin{figure}[t!]
    \centering
    \includegraphics[width=\linewidth]{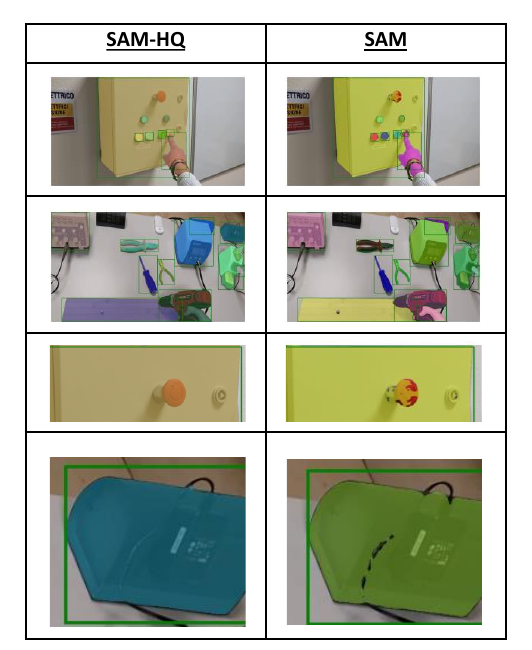}
    \caption{Comparison between SAM-HQ (left) and standard SAM (right). We reported also the bounding boxes (in green) used to generate segmentation masks.}
    \label{fig:sam_vs_hq}
\end{figure}

SAM-HQ~\cite{sam_hq} is an advanced extension of the Segment Anything Model (SAM~\cite{kirillov2023segany}), designed to enhance the segmentation of complex objects. SAM originally offered impressive scaling and zero-shot capabilities, but its mask prediction quality fell short, especially with intricate structures. To address this limitation, the authors of~\cite{sam_hq} proposed HQ-SAM, which retains SAM's promptable design, efficiency, and zero-shot generalizability while accurately segmenting any object. Considering the challenging nature of the industrial objects of the ENIGMA-51, we opted to use SAM-HQ as it proves to be a suitable solution for accurate segmentation. Figure~\ref{fig:sam_vs_hq} shows a comparison between SAM and SAM-HQ showing the better accuracy of the segmentation masks generated by SAM-HQ for wires and small buttons.  \\
\textbf{Implementation details:}
For the mask extraction, we used the SAM-HQ code provided in the official repository\footnote{\url{https://github.com/SysCV/sam-hq}}. We used the bounding-box annotations from the ENIGMA-51 dataset to prompt SAM-HQ, which enabled the generation of the desired masks. The checkpoint file ``sam\_hq\_vit\_h.pth", pretrained on HQSeg-44K~\cite{sam_hq}, was used for the model. During the inference phase, SAM-HQ generated a total of 55,427 hand masks and 270,519 object masks. The inference process required 6 hours using an NVIDIA A30 GPU. The semantic masks have been organized in structured JSON files, and they are released with the ENIGMA-51 dataset.

\subsubsection{Hand keypoints using MMPose}
\begin{figure}[t!]
  \centering
    \includegraphics[width=0.8\linewidth]{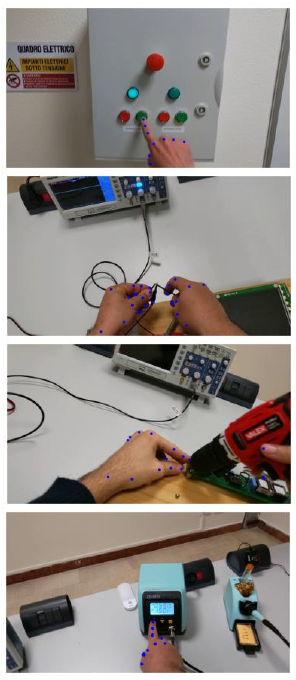}
    \caption{Hand Keypoints extracted with MMPose.}
    \label{fig:hand_keypoints}
\end{figure}

Since the hands represent the channel with which humans interact with the objects, we extracted hand keypoints using the MMPose~\cite{mmpose2020} framework with the aim of releasing pseudo-labels useful to study human-object interactions with the proposed ENIGMA-51 dataset.
MMPose~\cite{mmpose2020} is a useful open-source toolbox based on PyTorch, serving as part of the OpenMMLab project able to simultaneously detect the hands and localize their 2D keypoints.\\
\textbf{Implementation detail:}
We used the code provided in the official repository\footnote{\url{https://github.com/open-mmlab/mmpose}}. Since MMPose requires an input hand box, we used our hand annotations. We employed the pre-trained “onehand10k" model, which has been trained on images belonging to the Onehand10K\cite{onehand10k_wang} dataset with a resolution of 256x256. The model outputs keypoints for each hand, and each keypoint is associated with a confidence score ranging from 0 to 1. The confidence score allows us to filter out keypoints with lower accuracy. We saved all the extracted information in a JSON file. In total, we processed 30,747 left-hand bounding boxes and 24,680 right-hand bounding boxes using the MMPose framework. Figure~\ref{fig:hand_keypoints} shows some examples of 2D hand keypoints extracted with MMPose.

\subsubsection{Features extraction using DINOv2}
DINOv2~\cite{oquab2023dinov2} is a family of foundation models that produce universal features suitable for both image-level visual tasks (such as image classification, instance retrieval, and video understanding) and pixel-level visual tasks (including depth estimation and semantic segmentation).\\
\textbf{Implementation detail:} We used the official implementation~\footnote{\url{https://github.com/facebookresearch/dinov2}} with the publicly available dinov2\_vitg14 pre-trained model. Image preprocessing involved a transformation pipeline consisting of resizing and centre cropping the images to a resolution of 224x224, followed by converting them to tensors and applying normalization with mean and standard deviation values of ImageNet. Each frame was then processed using the model, obtaining a tensor of size (1, 1536). The output tensors representing the extracted features were saved in \textit{.npy} format and they will be released with the ENIGMA-51 dataset.

\subsubsection{Features extraction using CLIP}
To provide a set of features allowing further analysis and the study of downstream tasks with the ENIGMA-51 dataset, we exploited CLIP~\cite{Radford2021LearningTV} to extract text-image representations. We also used these features to explore human-object interactions with foundational models trained with generic and diverse data
and without domain-specific data.
CLIP~\cite{Radford2021LearningTV} is an advanced method for image representation learning from natural language supervision. It involves joint training of image and text encoders to predict correct pairings of (image, text) training examples. CLIP's architecture includes a simplified version of ConVIRT~\cite{zhang2022contrastive} trained from scratch, allowing for efficient and effective image representation learning.\\
\textbf{Implementation details:} We used the public implementation available to the following GitHub repository: \url{https://github.com/moein-shariatnia/OpenAI-CLIP}. The pretrained \textit{ViT-L/14@336px} model has been used, and the images were processed through a specific preprocessing step composed of resizing, center cropping, tensor transformation, and normalization. The output of CLIP is a tensor of size (1, 768) which is saved in \textit{.npy} format. All the extracted features will be released with the ENIGMA-51 dataset.

\section{Benchmark and Baselines Details}
\label{experiment}
\subsection{Untrimmed Temporal Detection of Human-Object Interactions}
Starting from the manually labeled timestamp of a key interaction, we defined the ground truth interaction temporal boundaries to employ our baseline based on ActionFormer~\cite{zhang2022actionformer}.
We tested two different strategies to set the interaction temporal boundaries.
The first consists of setting the start and end boundaries 15 frames before and after the labeled timestamp, respectively. Instead, in the second approach we set the action start at labeled interaction timestamp and we determined the action end empirically, allowing the model to observe hand movements after the labeled interaction timestamp for “take” and “release” actions. For “first contact” and “de-contact” interactions, the action end time was set 15 frames prior to the annotated timestamp.

The second approach demonstrated a better consistency in comparison to the first, despite yielding comparable mp-mAP scores during evaluations of “take” and “release” interactions ($41.45\%$ versus $42.27\%$). In particular, when applying a temporal threshold of 1 second, the second strategy yielded a p-mAP of $27.40\%$, distinctly outperforming the $11.25\%$ achieved by the first. This observation highlights that although both methods produce comparable results, the second approach has an advantage in setting the action at the labeled timestamp. This ensures reliable performance even at lower time thresholds.
\\
\textbf{Implementation Details:}
We have used a Two-Stream (TS) network~\cite{wang2018temporal} to extract video features. Each video chunk is set to a size of $6$, and there is no overlapping between adjacent chunks. With a video frame rate of $30$, we get $5$ chunks per second. For appearance features, we extract data from the Flatten 673 layer of ResNet-200~\cite{he2016deep} from the central frame of each chunk. Motion features are extracted from the global pool layer of BN-Inception~\cite{ioffe2015batch} from optical flow fields computed from the $6$ consecutive frames within each chunk. Motion and appearance features are then concatenated. We used models pre-trained on ActivityNet to extract these feature vectors\footnote{\url{https://github.com/yjxiong/anet2016-cuhk}.}.

We tested different numbers of levels of feature pyramid and different regression ranges. We found reliable results when using $3$ levels of the feature pyramid, respectively, with a regression range of [0, 2], [2, 5], [5, 10000]. We trained the model for $60$ epochs using a learning rate of $0.0001$, $5$ warmup epochs, and a weight decay of  $0.05$, following a cosine scheduler. All the experiments were conducted using $4$ Nvidia A30 graphics cards.

\subsection{Egocentric Human-Object Interaction Detection}
\begin{figure}[t!]
    \centering
    \includegraphics[width=0.5\columnwidth]{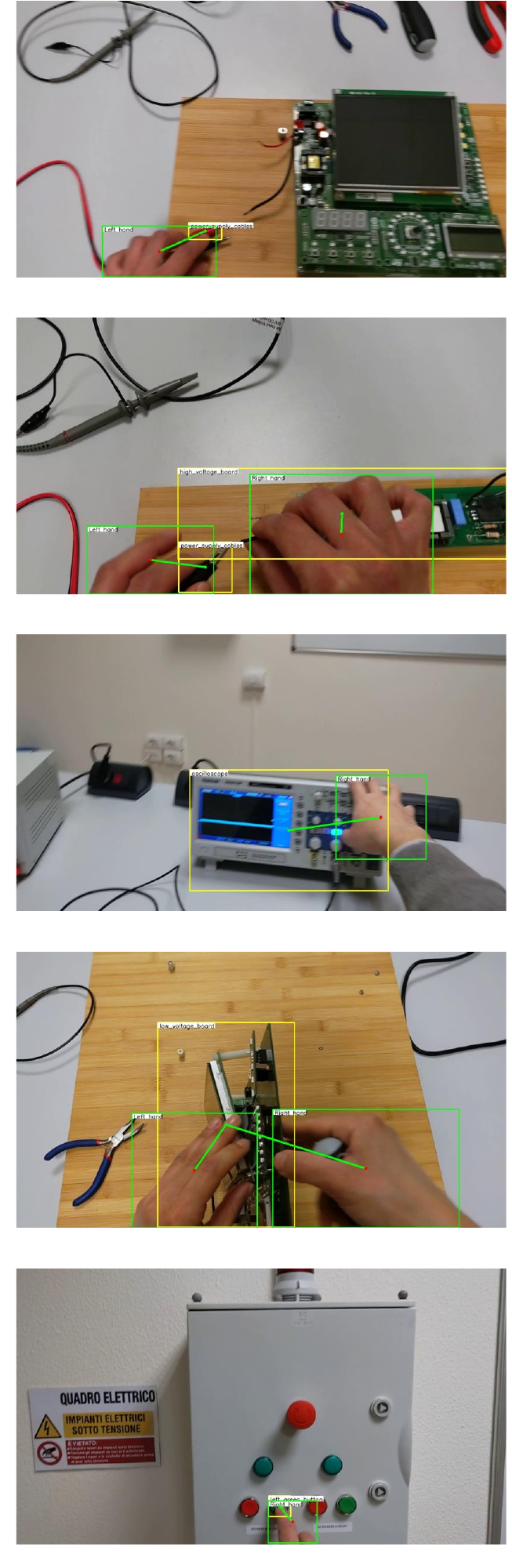}
  \caption{Qualitative results of the adopted baseline for the EHOI detection task.}
  \label{fig:qualitative_examples_ehoi}
\end{figure}

To perform the experiments for the EHOI detection task using the baseline based on \cite{leonardi2023exploiting}, we used a machine with a single \textit{NVIDIA A30} GPU and an \textit{Intel Xeon Silver 4310} CPU. We scaled all the images to a resolution of 1280x720 pixels. We trained the model using the \textit{Stochastic Gradient Descent} (SGD) for 80,000 iterations, an initial learning rate of 0.001, which is decreased by a factor of 10 after 40,000 and 60,000 iterations, and a minibatch size of 4 images. Figure~\ref{fig:qualitative_examples_ehoi} shows qualitative results of the adopted baseline. These qualitative results provide insights into the importance of incorporating domain-specific data during the training phase to extract objects knowledge useful to provide services to workers in the industrial domain.

\subsection{Short-Term Object Interaction Anticipation}
\begin{figure*}[ht!]
    \centering
    \includegraphics[width=\linewidth]{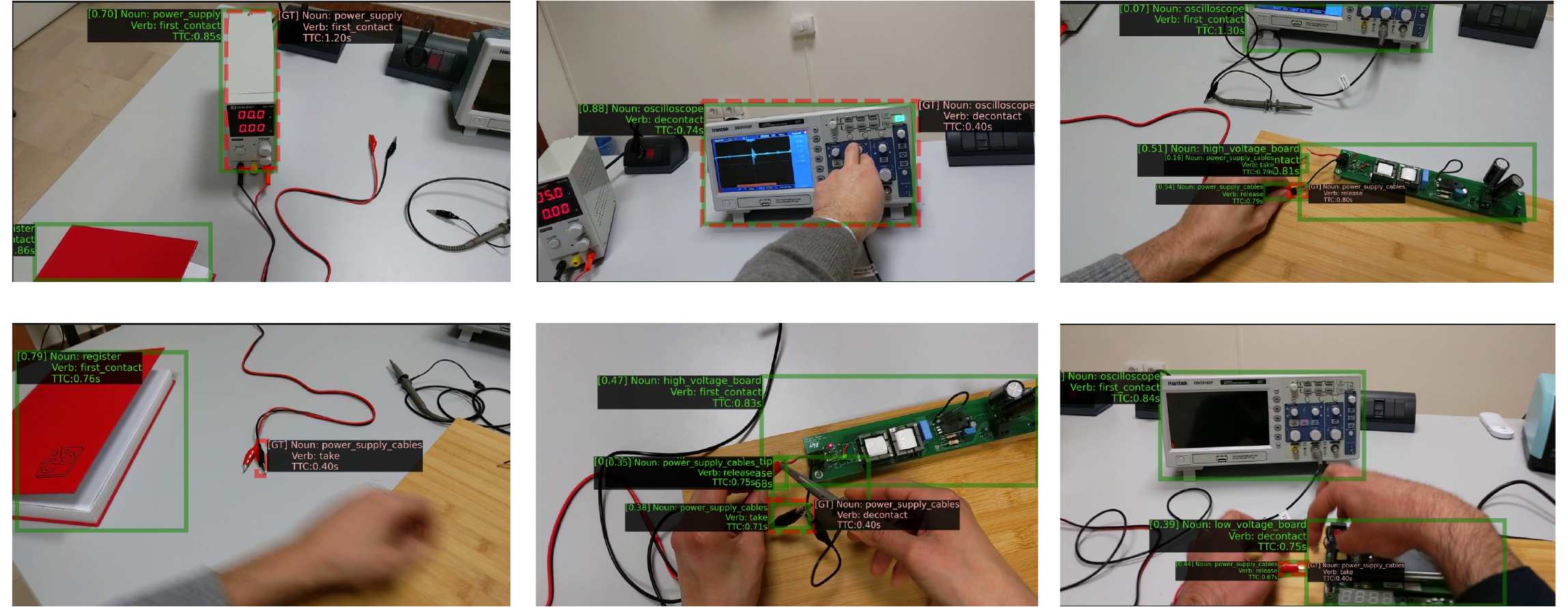}
    \caption{We reported qualitative results of our baseline based on StillFast\cite{ragusa2023stillfast} for the short-term object interaction anticipation task.}
    \label{fig:sta}
\end{figure*}

We achieve the short-term object interaction anticipation task with our baseline based on \cite{ragusa2023stillfast}. Figure \ref{fig:sta} shows some qualitative results. In the first row we reported correct predictions, while in the second one we reported wrong predictions. The predictions are represented with the green bounding boxes reporting the score, the noun and verb classes and the TTC, while the ground truth is shown in red with the name and verb classes and the TTC.
\\
\textbf{Implementation Details:} At training time, to obtain high-resolution images and low resolution videos, we used the same parameters used in \cite{ragusa2023stillfast}. At test time, we feed to the networks still images of height $H$ = 800 pixels and videos of height $h$ = 256 pixels.
The 2D backbone of the still branch is a ResNet-50 architecture. The weights of this backbone and the ones of the standard feature pyramid layer are initialized from a Faster R-CNN model\cite{ross2015faster} pre-trained on the COCO dataset \cite{lin2014COCO}. The 3D network which composes the fast branch is an X3D-M model \cite{feichtenhofer2020x3d} pre-trained on Kinetics \cite{carreira2017kineticsdataset}. 
The model has been trained with a base learning rate
of 0.001 and a weight decay of 0.0001. The learning rate
is lowered by a factor of 10 after 15 and 30 epochs. The
model is trained in half precision on four NVIDIA V100
GPUs with a batch size of 32.

\subsection{Natural Language Understanding of Intents and Entities}
We split our real dataset using an 80:20 ratio, with an identical test split employed across all experiments, which is uniformly distributed across intent and entity classes.  

DIETClassifier~\cite{bunk2020diet} was adopted for intent and entity prediction. The model has been trained on an Intel Core i5 CPU for 100 epochs with a learning rate of 0.001 and a variable batch size which linearly increases for each epoch from 64 to 256. 

\begin{figure*}[ht!]
    \centering
    \includegraphics[width=\linewidth]{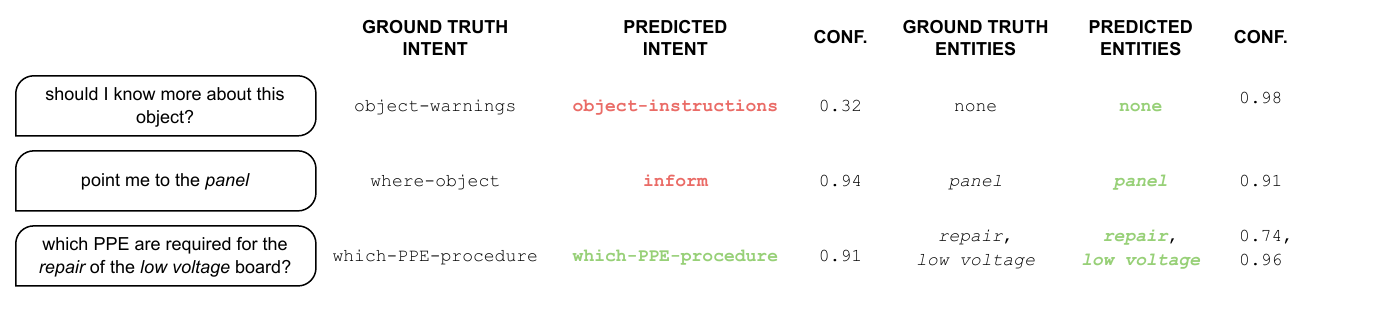}
    \caption{Qualitative results showing two incorrect intent predictions (first two rows) and a correct prediction (last row), alongside correct entity predictions.}
    \label{fig:iequalitative}
\end{figure*}

\begin{table*}[ht!]
   \centering
    \begin{tabular}{ccccccccc}
    \textbf{} & \multicolumn{4}{c}{\textbf{Intent}} & \multicolumn{4}{c}{\textbf{Entity}} \\ \cline{1-9} 
    \textbf{Training} & \multicolumn{1}{l}{\textbf{Accuracy}} & \textbf{Precision} & \textbf{Recall} & \textbf{F1-score} & \multicolumn{1}{l}{\textbf{Accuracy}} & \textbf{Precision} & \textbf{Recall} & \textbf{F1-score} \\ \hline
    real & \textbf{0.867} & \textbf{0.840} & \textbf{0.867} & \textbf{0.844} & 0.994 & 0.965 & 1.0 & 0.981 \\ 
    real+G10  & 0.830 & 0.822 & 0.830 & 0.815 & 1.0 & 1.0 & 1.0 & 1.0 \\ 
    real+G50  & 0.792 & 0.788 & 0.792 & 0.773 & 1.0 & 1.0 & 1.0 & 1.0 \\ 
    real+G100 & 0.792 & 0.794 & 0.792 & 0.784 & 1.0 & 1.0 & 1.0 & 1.0 \\
    G100 & 0.584 & 0.622 & 0.584 & 0.564 & \textbf{1.0} & \textbf{1.0} & \textbf{1.0} & \textbf{1.0}\\ \hline
    \end{tabular}
     \caption{Results for intents and entities classification considering different sets of training data.}
    \label{tab:fullintententity}
\end{table*} 

Table~\ref{tab:fullintententity} reports the results for the intent classification task (first four columns) and for the entity classification task (last four columns). Five different variations of the training set were explored: real data, real data + G10 data, real data + G50 data, real data + G100 data, and G100; and four different metrics were used for both intent and entity classification task: accuracy, precision, recall, and F1-score. The best results for the intent classification task have been obtained using only real data for training, with an accuracy, precision, recall, and F1-score of 0.867, 0.840, 0.867, and 0.844 respectively. However, using only generated data (G100) for training leads to poorer performances, with an accuracy, precision, recall, and F1-score of 0.584 (-0.283), 0.622 (-0.218), 0.584 (-0.283), 0.564 (-0.280) respectively. These results, in conjunction with the difficulties encountered during the prompting process, suggest that generated data does not fully reflect real utterances, and modern generative models may not accommodate all the constraints imposed by our specific context, thus affecting the model's performance. 
Exploring the use of combinations of real and generated data for training, we observe the best performances when the generated data is not predominant over the real data. In fact, we obtained an accuracy, precision, recall, F1-score of 0.830 (-0.037), 0.822 (-0.018), 0.830 (-0.037), and 0.815 (-0.029) respectively using real data and G10 for training, an accuracy, precision, recall, F1-score of 0.792 (-0.075), 0.788 (-0.052), 0.792 (-0.075), 0.773 (-0.071) respectively using real data and G50 for training, and an accuracy, precision, recall, F1-score of 0.792 (-0.075), 0.794 (-0.046), 0.792 (-0.075), 0.784 (-0.060) respectively using real data and G100. The performance deteriorates significantly with the addition of G10 to real data, subsequently showing a gradual decline as more generated data is introduced, ultimately stabilizing as G100 is added.  

Regarding entity classification, the results show that generated data alone is able to represent entities as how they're encountered in the real setting. In fact, best results are attained using G100 alone for training with an accuracy, precision, recall, and F1-score of 1.0, 1.0, 1.0, 1.0 respectively. However, real data itself attains near-perfect performances, with an accuracy, precision, recall, and F1-score of 0.994 (-0.006), 0.965 (-0.035), 1.0 ($\pm$0), 0.981 (-0.019) respectively.

Figure~\ref{fig:iequalitative} presents qualitative results for three distinct utterances for both intent and entity classification tasks. 
In the top row, we observe an utterance with an incorrect intent prediction (``object-instructions'' instead of ``object-warnings'') with a confidence of 0.32. This utterance did not contain any entities, and the absence of entities was correctly predicted with a confidence of 0.98. These results suggest that our model exhibited uncertainty in determining the appropriate class for the utterance. This uncertainty could be attributed to the fact that both ``object-instructions'' and ``object-warnings'' often contain utterances formulated in a very similar manner. However,  the model's capabilities concerning entity classification tend to be highly accurate with a significant confidence score. 
Moving to the middle row, we encounter an utterance with an incorrect intent prediction (``inform'' instead of ``where-object'') with a confidence score of 0.94. This utterance contained an entity of the ``object'' type, and the presence and class of this entity were correctly predicted with a confidence of 0.91. These results suggest that our model exhibited a high level of confidence in classifying the utterance, despite its incorrect classification. This observation may indicate that this particular utterance shares significant similarities with those typically found in the ``where-object'' intent. Similarly, in this instance, the model's capabilities enable accurate entity classification with a high confidence score.
Lastly, the bottom row showcases an utterance with a correct intent prediction (``which-PPE-procedure'')   with a confidence score of 0.91. This utterance featured entities of both ``procedure'' and ``board'' types, and the presence and classes of these entities were correctly predicted with confidence scores of 0.74 and 0.96, respectively. These results suggest that our model exhibited a high level of confidence in classifying the utterance, and its classification was indeed correct. Ultimately, in this latter scenario as well, the model's capabilities enable accurate entity classification, resulting in a confidence score ranging from moderate to high. 

\clearpage
{\small
\bibliographystyle{ieee_fullname}
\balance
\bibliography{main}
}

\end{document}